\documentclass[journal,twocolumn]{IEEEtran}
\usepackage{cite}
\usepackage{amsmath,amssymb,amsfonts,bm}

\usepackage{algorithmic}
\usepackage{graphicx}
\usepackage{subfigure}
\usepackage{textcomp}
\usepackage{makecell}
\usepackage{booktabs}
\usepackage{multirow}
\usepackage{color}
\usepackage{bm}

\usepackage{amsthm}
\usepackage{algorithm}
\usepackage{algorithmic}

\ifCLASSINFOpdf
\else
\fi
\begin{document}
%
%
%
%
\title{FedBM: Stealing Knowledge from Pre-trained Language Models for Heterogeneous  Federated Learning}
\author{Meilu Zhu, Qiushi Yang, Zhifan Gao, Yixuan Yuan*,Jun Liu* 
\thanks{This work was supported by the Hong Kong Research Grants Council under Grant 11212321, 11217922, and ECS-21212720, the HKSAR Innovation and Technology Commission (ITC) under ITF Project MHP/109/19, ITS/229/22, and the Science, Technology and Innovation Committee of Shenzhen under Grant SGDX20210823104001011. (\textit{*Corresponding authors: Jun Liu (dr.jun.liu@hku.hk), Yixuan Yuan (yxyuan@ee.cuhk.edu.hk)})}
\thanks{
M. Zhu is with Department of Mechanical Engineering, City University of Hong Kong;
Q. Yang is with Department of Electrical Engineering, City University of Hong Kong;
Z. Gao is with School of Biomedical Engineering, Sun Yat-sen University;
Y. Yuan is with Department of Electronic Engineering, Chinese University of Hong Kong;
J. Liu is with Department of Department of Data and Systems Engineering, The University of Hong Kong.}}

\markboth{Journal of \LaTeX\ Class Files,~Vol.~14, No.~8, August~2015}%
{Shell \MakeLowercase{\textit{et al.}}: Bare Demo of IEEEtran.cls for IEEE Journals}
%



\maketitle

\begin{abstract}
Federated learning (FL) has shown great potential in medical image computing since it provides a decentralized learning paradigm that allows multiple clients to train a model collaboratively without privacy leakage. However, current studies have shown that data heterogeneity incurs local learning bias in classifiers and feature extractors of client models during local training, leading to the performance degradation of a federation system. 
To address these issues, we propose a novel framework called \underline{F}ederated \underline{B}ias eli\underline{M}inating (FedBM) to get rid of local learning bias in heterogeneous federated learning (FL), which mainly consists of two modules, \textit{i.e.}, Linguistic Knowledge-based Classifier Construction (LKCC) and Concept-guided Global Distribution Estimation (CGDE). Specifically, LKCC exploits class concepts, prompts and pre-trained language models (PLMs) to obtain concept embeddings. These embeddings are used to estimate the latent concept distribution of each class in the linguistic space. Based on the theoretical derivation, we can rely on these distributions to pre-construct a high-quality classifier for clients to achieve classification optimization, which is frozen to avoid classifier bias during local training. 
CGDE samples probabilistic concept embeddings from the latent concept distributions to learn a conditional generator to capture the input space of the global model. Three regularization terms are introduced to improve the quality and utility of the generator. The generator is shared by all clients and produces pseudo data to calibrate updates of local feature extractors.
Extensive comparison experiments and ablation studies on public datasets demonstrate the superior performance of FedBM over state-of-the-arts and confirm the effectiveness of each module, respectively. The code is
 available at https://github.com/CUHK-AIM-Group/FedBM.
\end{abstract}

\section{Introduction}
\label{sec:introduction}
With the explosive growth of data, training deep models has become a promising path to achieve high-precision computer-aided diagnosis (CAD)~\cite{razzak2018deep,yan2018joint,zhu2021dsi,zhu2024DEeR}. However, centralizing data from different hospitals or institutions to construct large-scale medical training datasets is unrealistic, thanks to growing privacy concerns or legal restrictions~\cite{FedOSS}. To conquer this problem, a new training paradigm, federated learning (FL)~\cite{fedavg,fedprox,zhufeddm2023}, is proposed to learn deep models across different clients (hospitals) under the coordination of a cloud server. In each round of FL training, clients independently train local models on their private data and upload them to the server, where the models are aggregated. The aggregated model is then sent back to the clients, serving as the initialization for the next training round. Importantly, clients are not required to share their raw data during this process, thereby preserving their privacy. 
\begin{figure}[!t]
\centering
\includegraphics[width = 0.49\textwidth]{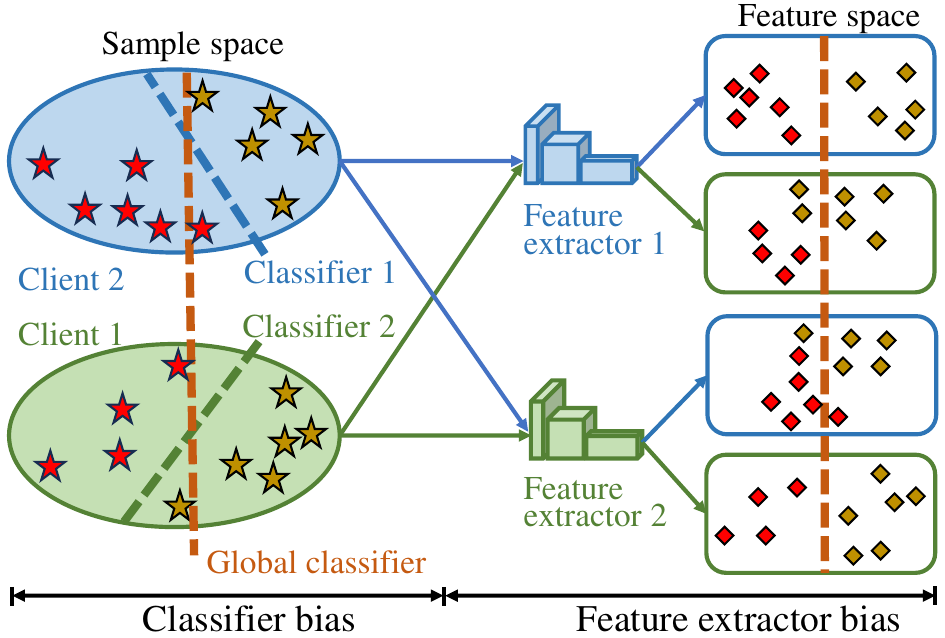}
\caption{Data heterogeneity causes local learning bias, including classifier bias and feature extractor bias. (Best viewed in color)}
\label{fig:challenge}
\end{figure}
Unfortunately, the heterogeneity among client datasets significantly contributes to \textit{local learning bias} at the client side, leading to performance degradation in federation systems~\cite{li2021fedbn, fedprox,jeong2018communication, guo2023fedbr}. The learning bias primarily manifests in two aspects from the perspective of representation learning, as shown in Fig.~\ref{fig:challenge}. 

Firstly, local learning bias appears in classifiers of local models during training \cite{luo2021no,xu2023personalized}. When data from clients are heterogeneous, the local classifiers are dominated by their local class distributions, leading to the shifted decision boundaries across clients in Fig.~\ref{fig:challenge}. Recent studies~\cite{huang2023rethinking, dai2023tackling, tan2022fedproto, qi2023cross, long2023fedcd} have attempted to exploit class prototypes as classifier to avoid this problem. However, these approaches obtain limited performance since the quality of class prototypes for one client is affected by its biased local feature extractor. In experiments, we surprisingly found that a simple strategy of using a fixed randomly-initialized classifier for all clients outperforms the baseline method, FedAvg~\cite{fedavg}, as shown in Table~\ref{tab:ablation_prompt_method}. The results indicate that sharing a fixed classifier across clients is a feasible path to alleviate the classifier bias problem. Intuitively, random initialization is not the optimal strategy to build the fixed classifier since it does not consider intra-class semantic information and inter-class distance relations. This inspires us to explore a better solution to pre-construct a high-quality classifier for clients and freeze it during federated training. 

Secondly, the heterogeneous data of clients would produce inconsistent local feature extractors. The features from the feature extractor of one client may differ significantly from those extracted by the feature extractor of another client, even for the same input data in Fig.~\ref{fig:challenge}. Consequently, the global feature extractor, obtained by aggregating these inconsistent local feature extractors, will fail to extract generalizable features for adapting to all clients~\cite{guo2023out}. Previous works~\cite{fedprox,li2021model,acar2021federated, karimireddy2020scaffold} reduce the inconsistency by regularizing the distance between local models of the current round and the global model from the last round. However, it is hard to balance the trade-offs between optimization and regularization to perform well~\cite{wang2020tackling}. Different from these methods, we try to directly narrow the gap between the data distribution of clients by exploiting textual prior to estimate global distribution to supplement the local distribution.

To tackle the aforementioned issues, we propose a novel framework called \underline{F}ederated \underline{B}ias eli\underline{M}inating (FedBM) to get rid of local learning bias in heterogeneous federated learning (FL). FedBM mainly consists of two modules, \textit{i.e.}, Linguistic Knowledge-based Classifier Construction (LKCC) and Concept-guided Global Distribution Estimation (CGDE). Specifically, LKCC exploits class concepts, prompts, and pre-trained language models (PLMs) to obtain concept embeddings. These embeddings are used to estimate the latent concept distribution of each class in the linguistic space. Based on the theoretical derivation, we can rely on these distributions to construct a high-quality global classifier for alignment between visual and linguistic spaces, avoiding classifier bias. CGDE samples probabilistic concept embeddings from the latent concept distributions to learn a conditional generator to capture the input space of the global model. 
Three regularization terms are introduced to improve the quality and utility of the generator. The generator is shared by all clients and produces pseudo data to calibrate updates of local feature extractors. 
The contributions of this work are summarized as follows:
\begin{itemize}
\item We present a novel \underline{F}ederated \underline{B}ias eli\underline{M}inating (FedBM) framework, which represents the first effort to use linguistic knowledge to address heterogeneous FL.

\item We propose Linguistic Knowledge-based Classifier Construction (LKCC) that exploits linguistic knowledge from pre-trained language models (PLMs) to pre-define a high-quality global classifier to avoid classifier bias.

\item We design Concept-guided Global Distribution Estimation (CGDE) that utilizes probabilistic concept embeddings to learn a conditional generator to produce pseudo data to calibrate updates of local feature extractors.

\item We conduct extensive experiments on public datasets to evaluate the proposed framework. The results show the superior performance of FedBM against state-of-the-arts and the effectiveness of different modules.
\end{itemize}

This work builds upon our conference paper \cite{Zhu_Stealing_MICCAI2024} and extends it in the following aspects:
(1) It comprehensively discusses the local learning bias problem from the perspectives of the classifier and feature extractor of a local model. In addition to debiasing local classifiers as in the conference version, it proposes a novel CGDE module to eliminate the learning bias of local feature extractors, thereby achieving robust local training; (2) It provides an exhaustive review of existing methods focusing on the local learning bias problem; (3) It introduces three new datasets to further verify the effectiveness and generalization of the proposed method across various medical tasks; (4) The experimental results show that this extended version achieves better performance than the conference version, with significant improvements; (5) More comprehensive ablation experiments are conducted to verify the effectiveness of different modules of the proposed method and its scalability to different numbers of clients.

\textbf{Roadmap.} The rest of the paper is organized as follows. In Section \ref{sec:related-work}, we review previous methods focusing on classifier debiasing, model alignment and data augmentation in FL. In Section \ref{sec:method}, the proposed FedBM is introduced in detail. We describe the implementation details and verify of the effectiveness of the proposed FedBM in Section \ref{sec:experiments}. Finally, the paper is closed with the conclusion in Section \ref{sec:conclusion}.

\section{Related Work}
\label{sec:related-work}
We introduce existing methods of classifier debiasing, model alignment and data augmentation in federated learning.

\subsection{Classifier Debiasing in Federated Learning}
Federated learning (FL) provides a new solution to handle privacy concerns in distributed training. 
As the pioneering method, FedAvg~\cite{fedavg}, trains a global model by aggregating local models from multiple clients without accessing their raw data. However, it undergoes considerable performance degradation when the data of clients are heterogeneous due to various imaging protocols, disease incidences, or population demographics. One of the main reasons is that data heterogeneity results in divergent local classifiers. Current approaches to this problem can be divided into three categories.

The first type of approaches~\cite{luo2021no,huang2024federated,zhou2023fedfa} aims to generate a balanced feature set to train local classifiers. For example, CCVR~\cite{luo2021no} exploits feature representations of all clients to build an approximated Gaussian mixture model, which is sent to each client for sampling more virtual representations. RUCR~\cite{huang2024federated} broadcasts global prototypes to clients and arbitrarily fuses them and local features to synthesize virtual features.
The second category~\cite{huang2023rethinking, dai2023tackling, tan2022fedproto, qi2023cross, long2023fedcd} tends to replace local classifiers with class prototypes. For instance, FedProto~\cite{tan2022fedproto} directly aggregates prototypes of each class as local classifiers of clients. FedNH~\cite{dai2023tackling} produces uniformly-distributed class prototypes as initial local classifiers, and then smoothly infuses the class semantics into class prototypes. FPL~\cite{huang2023rethinking} uploads feature representations of all clients to the server and clusters them to get different prototype centers for each class. These prototype centers are further averaged as local classifiers.

The third branch of works~\cite{fedETF, zhang2024upload} pre-constructs a fixed classifier before federated training. According to the theory of neural collapse~\cite{papyan2020prevalence} that classifier vectors converge to an optimal simplex equiangular tight frame (ETF) when the dataset is balanced and sufficient, FedETF~\cite{fedETF} and FedKTL~\cite{zhang2024upload} introduce a synthetic simplex ETF as a fixed classifier for all clients. However, the orthogonal relation between classifier vectors is too strict and lacks of semantic interpretability. In this work, we propose to borrow linguistic knowledge from pre-trained language models to construct local classifiers.  

\subsection{Model Alignment in Federated Learning}
Data heterogeneity also leads to misalignment between client models, i.e., client-level variance,  resulting in unstable and slow convergence during federated learning~\cite{li2020federated}. 
FedProx~\cite{fedprox} is the first work to solve this problem by introducing a proximal term into the objective during local training to restrict the distance between the current global model and the local model.  SCAFFOLD~\cite{karimireddy2020scaffold} introduces control variates to correct the drift in local updates. Nevertheless, the direct constraint in the parameter space may negatively affect model learning. 

Apart from the above solutions, another way is to introduce the constraint in the feature space to solve this problem. For example,
MOON~\cite{li2021model} presents model-level contrastive learning to maximize the similarity between the features of local models in the current round and the global model and minimize the similarity between the features of local models of the current round and the previous round.  
FedFA~\cite{zhou2023fedfa}, FedFM~\cite{ye2023fedfm}, and FedPAC~\cite{xu2023personalized} collect local class prototypes to generate global prototypes. These global prototypes are sent to clients as the alignment objective of feature representations during local training. 
RUCR~\cite{huang2024federated} pulls features within the same class towards corresponding global prototypes and
pushes features of the other classes away.
Although these prototype-based methods can improve representation learning, their performance highly relies on the quality of global prototypes. Unlike the existing methods, we directly narrow the distribution gap between client data.
\begin{figure*}[!t]
\centering
\includegraphics[width = 1.0\textwidth]{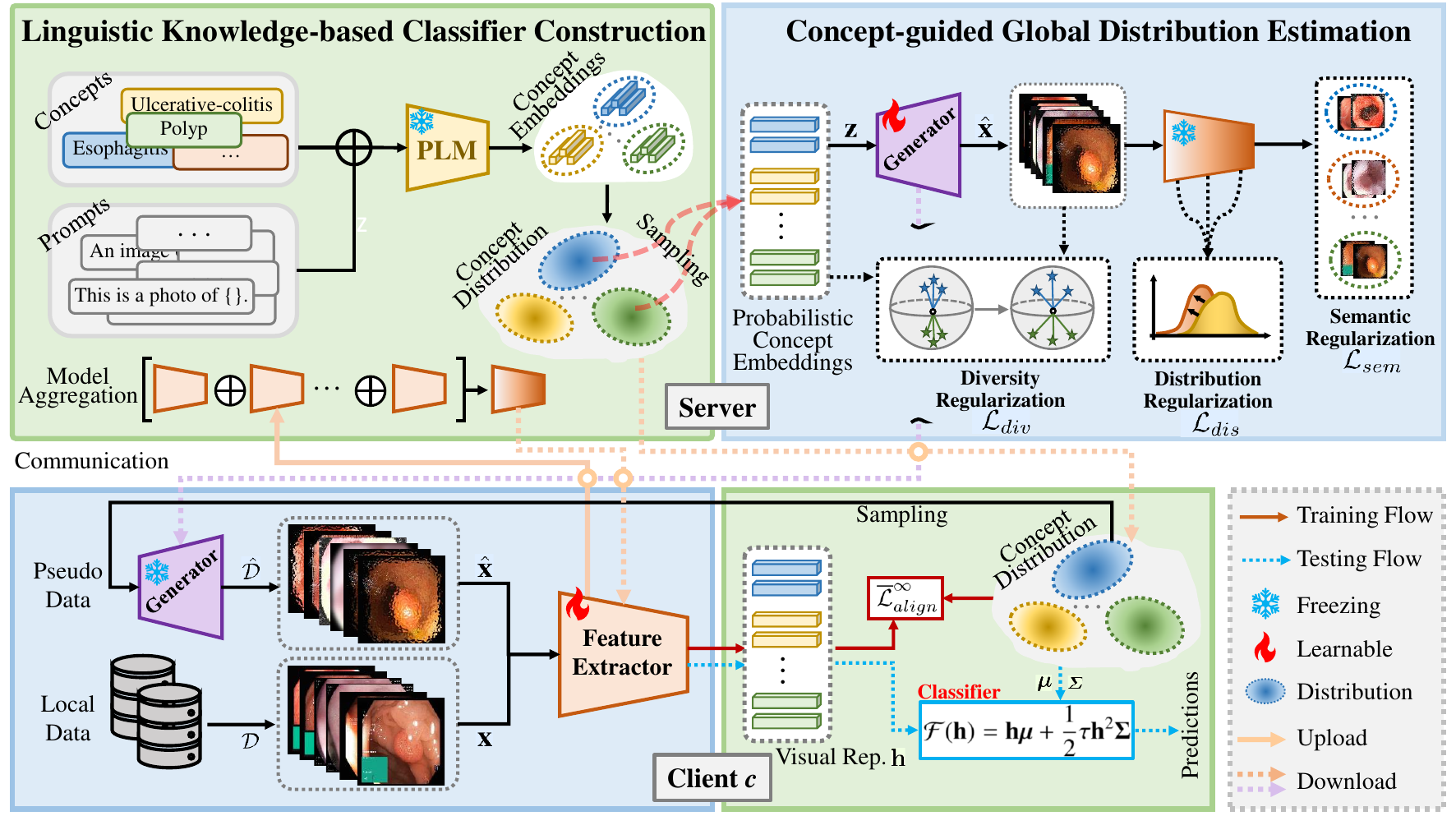}
\caption{The overview of the proposed FedBM framework. FedBM contains Linguistic Knowledge-based Classifier Construction (LKCC) and Concept-guided Global Distribution Estimation (CGDE). LKCC uses class concepts, prompts and PLMs to build latent concept distributions, which are sent to clients as local classifiers. CGDE samples probabilistic concept embeddings from the distributions to train a conditional generator. The generator is shared by all clients and produces pseudo data to calibrate updates of local feature extractors. (Best viewed in color) }
\label{framework}
\end{figure*}

\subsection{Data Augmentation in Federated Learning}
Data Augmentation is a commonly-used way to relieve data heterogeneity issues in federated learning. 
\cite{de2022mitigating} have verified that some common data augmentation techniques can significantly improve out-of-domain generalization in federated settings, such as random cropping,  horizontal flipping and color transformations. Besides, Mixup~\cite{zhang2017mixup} also obtains the widespread attention~\cite{guo2023fedbr, yoon2021fedmix, shin2020xor}. For example, FedMix~\cite{yoon2021fedmix} averages local batches to produce synthetic data. The server gathers these data and then sends them to clients. Clients combine these synthetic data with their local data to perform Mixup in local training.

In addition, FedOV~\cite{diao2023towards} and FedOSS~\cite{FedOSS} are inspired by adversarial training and use fast gradient sign method (FGSM) to generate unknown samples.
Moreover, FedRDN~\cite{yan2023simple} and CCVR~\cite{luo2021no} hypothesize that the images of each client are sampled from a multivariate Gaussian distribution. By sharing Gaussian distributions across clients, each client can sample augmented data to enhance local data, thereby reducing the domain gap.
SDA-FL~\cite{li2022federated} pre-trains a generative adversarial network (GAN)~\cite{goodfellow2014generative} to generate synthetic data in each client. These synthetic data are then collected by the server to construct a global synthetic dataset to optimize global model. 
FedDiff~\cite{mendieta2024navigating} trains a class-conditioned diffusion model~\cite{ho2020denoising} on local data at the clients. These local diffusion models are sent to the server to generate data for training global model.

Data-free knowledge distillation is also a popular way to synthesize samples in the federated setting~\cite{zhang2022fine, zhang2022dense, wang2024dfrd}. For instance, DENSE~\cite{zhang2022dense} and FedFTG~\cite{zhang2022fine} utilize the ensemble client models to train a generator and then generate synthetic data to train global model at the server. However, these methods learn a generator to capture the mapping between random noises (from Gaussian distribution) and samples in the image space. The random noises lack semantically meaningful information and do not form well-organized class clusters, enabling the generator to produce low-quality images. The proposed FedBM framework introduces the text information of classes to remedy this disadvantage.

\section{Method}
\label{sec:method}
This section first presents the workflow of FedBM and then introduces its submodules as well as optimization process.

\subsection{Overview of FedBM}
We present a \underline{F}ederated \underline{B}ias eli\underline{M}inating (FedBM) framework to remove local learning bias in heterogeneous federated learning. FedBM follows a standard FL training paradigm and consists of $C$ distributed clients and a trustworthy server. Each client possesses a local dataset $\mathcal{D}^c = \{(\mathbf{x}_i^c,  \mathbf{y}_i^c)\}_{i=1}^{N_c}$ with $K$ classes, where ${N_c}$ is the sample number of $\mathcal{D}^c$, and $\mathbf{x}_i^c$ is a training instance with the label $\mathbf{y}_i^c$. The goal of FedBM is to coordinate these clients to train a global model $\mathcal{F}(\textrm{W}_{fe}, \textrm{W}_{fc})$, where $\mathcal{F}$ contains a feature extractor $\mathcal{F}(\textrm{W}_{fe})$ and a feature classifier $\mathcal{F}(\textrm{W}_{fc})$. The overall training process proceeds through communication between clients and the server for multiple rounds. Concretely, we first pre-construct a high-quality global classifier via Linguistic Knowledge-based Classifier Construction (LKCC) before distributed training. Next, the $c$-th client downloads the global feature extractor and classifier from the server to initialize the weights of its local model. During local training, all clients freeze local classifiers to avoid the classifier bias problem and only train their feature extractors. After local training, the local feature extractors $\mathcal{F}_c(\textrm{W}_{fe}^c)$ of clients are uploaded to the server to update the global feature extractor via model aggregation: $\mathcal{F}=\frac{1}{C}\sum_{c\in [C]} \mathcal{F}_c(\textrm{W}_{fe}^c)$. Concept-guided Global Distribution Estimation (CGDE) exploits the  aggregated global model and concept prior to train a conditional generator that can capture the input space of the global model. The global feature extractor is sent to each client as the initialization of the next round. The generator is also distributed to per client to produce domain-invariant samples to regularize local updates in a consistent direction. The overall framework is shown in Fig.~\ref{framework}.

\subsection{Linguistic Knowledge-based Classifier Construction}
Although pre-constructing a global classifier is a flexible way to mitigate local learning bias in classifiers of client models, it is very challenging since we do not have any knowledge about a high-quality classifier. Motivated by recent language-to-vision models \cite{CLIP,BiomedCLIP},
natural language descriptions (such as diagnosis reports) carry rich semantic information and can represent diverse images or clinical scans of different categories.  
Based on this insight, we propose Linguistic Knowledge-based Classifier Construction (LKCC), which exploits linguistic knowledge from pre-trained language models to construct a high-quality classifier for all clients.

As shown in Fig.~\ref{framework}, the server first collects a concept set $\{P_k\}_{k=1}^{K}$ from clients, where $P_k$ is the class name of the $k$-th class and $K$ is the total class number.
A set of $M$ predetermined prompts (such as ``This is an image of \{concept\}'' and ``The image shows \{concept\}.'' and so on) is used to contextualize the concepts. We input the contextualized concepts into a pre-trained language model (PLM) (such as the text encoder of BiomedCLIP~\cite{BiomedCLIP}) to obtain a set of concept embeddings $\bm{E}$, where $\bm{E} = \cup_{k=1}^{K}\{\bm{e}_1^{(k)}, \bm{e}_2^{(k)},...,\bm{e}_M^{(k)} \}$. PLM is trained on large-scale datasets based on contrastive learning and demonstrates strong feature transferability. Therefore, the obtained concept embeddings in $\bm{E}$ contain rich semantics, which has two favorable properties: (i) the distance relationship between concepts can be reflected through their similarities, (ii) concept embeddings in the linguistic space are domain-agnostic. Next, we employ these concept embeddings to build a high-quality classifier for clients. 

Given the concept embeddings of each class, a natural idea is to regard them as a multi-way classifier to train the local feature extractor by performing alignment between image representations and these embeddings, which can be formulated as minimizing the following contrastive loss:
\vspace{-2.0mm}
\begin{equation}
\footnotesize
\begin{aligned}
&\mathcal{L}_{align}= \\
&\frac{1}{N_c}\sum^{N_c}_{i=1} \mathbb{E}_{\bm{e}^{(\textbf{y}_i)} \in \bm{\Omega}^{(\textbf{y}_i)}}
\left( -\log \frac{e^{\tau\textbf{h}_i^{\textrm{T}}\bm{e}^{(\textbf{y}_i)}}}{e^{\tau\textbf{h}_i^{\textrm{T}}\bm{e}^{(\textbf{y}_i)}} +\sum^{K}_{k\neq \textbf{y}_i}
\mathbb{E}_{\bm{e}^{(k)} \in \bm{\Theta}^{(\textbf{y}_i)}}
e^{\tau\textbf{h}_i^{\textrm{T}}\bm{e}^{(k)}}} \right),
\label{eq1}
\end{aligned}
\end{equation}
where $\textbf{h}_i = \frac{\mathcal{F}(\textbf{x}_i)}{\|\mathcal{F}(\textbf{x}_i)\|_2}$ is the normalized representation of a sample $\textbf{x}_i$ of the client $c$. We add a fully-connected layer on top of the feature extractor  to align the dimension of $\textbf{h}_i$ and $\bm{e}_m^{(\textbf{y}_i)}$. $\bm{\Omega}^{(y_i)}$ is the positive embedding set of the class $\textbf{y}_i$ and contains the embeddings $\{\bm{e}_1^{(\textbf{y}_i)}, \bm{e}_2^{(\textbf{y}_i)},...,\bm{e}_M^{(\textbf{y}_i)} \}$. $\bm{\Theta}^{(\textbf{y}_i)}$ is the negative embedding set and contains the concept embeddings of the other categories. $\tau$ is the temperature coefficient. Generally, more diverse prompts can obtain richer concept embeddings (corresponding to language descriptions) to comprehensively describe one class. Aligning image representations to these concept embeddings can force the model to learn exhaustive visual details. Hence, the performance of the model $\mathcal{F}$ highly depends on the number $M$ of prompts under the supervision of Eq.~(\ref{eq1}).
However, it is difficult to obtain all prompts for a specific task via prompt engineering. Besides, different prompts should not be treated equally due to different importance. 
Considering these issues, we propose to further generalize Eq.~(\ref{eq1}) to the infinite space, namely, aligning the image representations and the concept embedding distribution of each class.

Assuming that the concept embeddings $\{\bm{e}_1^{(k)}, \bm{e}_2^{(k)},...,\bm{e}_M^{(k)} \}$ of the $k$-th class are sampled from a Gaussian distribution $\mathcal{N}(\bm{\mu}_k, \bm{\Sigma}_k)$, we compute the mean $\bm{\mu}_k$ and variance $\bm{\Sigma}_k$ as follows:
\begin{equation}
\small
\bm{\mu}_k = \frac{1}{M}\sum_{m=1}^{M} \bm{e}_m^{(k)},  \ \ \ \ \bm{\Sigma}_k = \frac{1}{M-1}\sum_{m=1}^{M} (\bm{e}_m^{(k)}-\bm{\mu}_k)(\bm{e}_m^{(k)}-\bm{\mu}_k).
\label{eq2}
\end{equation}
After estimating the distributions $\{\mathcal{N}^{(k)}\}_{k=1}^K$ of all classes, we can sample infinite concept embeddings, which correspond to instances with different characteristics in the image space. In the context, Eq.~(\ref{eq1}) can be reformulated as
\begin{equation}
\footnotesize
\begin{aligned}
&\mathcal{L}^\infty_{align}= \\
&\frac{1}{N_c}\sum^{N_c}_{i=1} \mathbb{E}_{\bm{e}^{(\textbf{y}_i)} \thicksim \mathcal{N}^{(\textbf{y}_i)}}
\left( -\log \frac{e^{\tau\textbf{h}_i\bm{e}^{(\textbf{y}_i)}}}{e^{\tau\textbf{h}_i\bm{e}^{(\textbf{y}_i)}} +\sum^{K}_{k\neq \textbf{y}_i}
\mathbb{E}_{\bm{e}^{(k)} \thicksim \mathcal{N}^{(k)}}
e^{\tau\textbf{h}_i\bm{e}^{(k)}}} \right).
\label{eq3}
\end{aligned}
\end{equation}
$\mathcal{L}^\infty_{align}$ is difficult to compute its exact form when the sampled concept embeddings are infinite. Here, we can derive its upper bound based on the existing method~\cite{wang2019implicit} and find a surrogate loss $\overline{\mathcal{L}}^\infty_{align}$:
\begin{equation}
\footnotesize
\begin{aligned}
\mathcal{L}^\infty_{align} \leq \overline{\mathcal{L}}^\infty_{align} = \frac{1}{N^c}\sum^{N_c}_{i=1} \left( -\log \frac{e^{\mathcal{F}(\textbf{h}_i,\textbf{y}_i)}}{\sum^K_{k=1}e^{\mathcal{F}(\textbf{h}_i, k)} } + \frac{\tau^2}{2}\textbf{h}_i^2\bm{\Sigma}_{(\textbf{y}_i)} \right),
\label{eq4}
\end{aligned}
\end{equation}
where $\mathcal{F}(\textbf{h}_i, k) = \textbf{h}_i\bm{\mu}_{(k)} + \frac{1}{2}\tau\textbf{h}_i^2\bm{\Sigma}_{(k)}$. The detailed derivation is shown in Appendix.
By minimizing the loss $\overline{\mathcal{L}}^\infty_{align}$, we can implement the alignment between the image representations and the concept embedding distributions.
It can be observed that $\overline{\mathcal{L}}^\infty_{align}$ is a softmax-based cross-entropy loss over $\mathcal{F}(\textbf{h}_i, k)$, with a constraint on variance of features. Therefore, we redefine the local classifier as
\begin{equation}
\mathcal{F}(\textbf{h}) = \textbf{h}\bm{\mu} + \frac{1}{2}\tau\textbf{h}^2\bm{\Sigma},
\label{eq5}
\end{equation}
where $\textbf{h} \in \mathbb{R}^{B\times D}$, $\bm{\mu} \in \mathbb{R}^{D\times K} $ and $\bm{\Sigma} \in \mathbb{R}^{D\times K}$. $B, D$ and $K$ are the batch size, the feature dimension and the class number. $\mathcal{F}(\textbf{h}) \in \mathbb{R}^{B\times K}$ denotes the logit outputs of the batch images. During the inference phase, the calculation of $\mathcal{F}(\textbf{h}) \in \mathbb{R}^{B\times K}$ do not rely on the class labels of $\textbf{h}$. Noticeably, the construction process is only conducted once and does not incur a high computation cost. Meanwhile, the constructed classifier only needs one round of transmission, which reduces communication overhead. 

Essentially, Eq.(\ref{eq1}) averages concept embeddings as the local classifier and treats all prompts equally. By comparison, the classifier $\mathcal{F}(\textbf{h})$ in Eq.~(\ref{eq5}) considers the variance of the concept embeddings and thus are more robust to match the semantic diversification of image representations, thereby achieving more accurate classification. Additionally, concept embedding distributions are derived from embeddings generated by a pre-trained language model (PLM). Since PLM is trained on large-scale datasets, these embeddings carry rich semantic information and can represent specific image samples. A concept embedding distribution can represent a specific cluster of samples, enabling the classification model to easily capture the relationship between concept embedding distributions and visual images. In contrast to FedETF~\cite{fedETF} that utilizes orthogonal initialization to construct the classifier, our method has stronger semantic interpretability and the elastic constraints on the inter-class angular margin.

\subsection{Concept-guided Global Distribution Estimation}
Data heterogeneity causes local client models to gradually forget the global knowledge learned in previous rounds during local training because they can only receive local data information and thus always optimize towards their own local distributions, resulting in inconsistency in local feature extractors $\mathcal{F}_c(\textrm{W}_{fe}^c)$. It has been confirmed that the inconsistency will incur sharper loss landscape and performance degradation of the global model~\cite{shi2023improving, chen2021personalized}. To break this dilemma, we propose Concept-guided Global Distribution Estimation (CGDE) to train a conditional generator based on the aggregated model and concept prior at the server, which can generate data that have a similar distribution to the input space of the global model. These generated data are combined with local data to train local feature extractors and restrict local updates, reducing local learning bias, as illustrated in Fig.~\ref{framework}.

Specifically, we consider a conditional generator $\mathcal{G}(\cdot)$ with the parameters $\textrm{W}_\mathcal{G}$, which takes a condition pair $(\textbf{z}_i, \hat{\textbf{y}}_i)$ as input to generate the synthetic sample $\hat{\textbf{x}}_i = \mathcal{G}(\textbf{z}_i, \textrm{W}_\mathcal{G})$. The label $\hat{\textbf{y}}_i$ is randomly sampled from a uniform distribution. 
Generally, synthetic data generated by a well-trained conditional generator should satisfy several key characteristics: \textbf{semantic similarity}, \textbf{data diversity}, \textbf{distribution consistency}, and \textbf{interpretability of conditions}.  Towards these goals, we introduce some regularization terms into the optimization objective to restrict the training of the generator $\mathcal{G}(\cdot)$ from these aspects, ensuring its quality and utility.

\textbf{Semantic Regularization} The generator is expected to produce synthetic samples with semantic similarity to instances in the input space of the global model. To put it differently, we hope that a generated sample $\hat{\textbf{x}}_i$ can be classified into the class $\hat{\textbf{y}}_i$ with a high probability by the global model. Therefore, we treat the global model $\mathcal{F}$ as the teacher model to optimize the generator $\mathcal{G}(\cdot)$ via the following loss:
\begin{equation}
\mathcal{L}_{sem} = CE( \mathcal{F}(\mathcal{G}(\textbf{z}_i, \textrm{W}_\mathcal{G}), \textrm{W}_{fe}, \textrm{W}_{fc}) ,\hat{\textbf{y}}_i),
\label{eq6}
\end{equation}
where $\mathcal{F}(\mathcal{G}(\textbf{z}_i, \textrm{W}_\mathcal{G}), \textrm{W}_{fe}, \textrm{W}_{fc})$ is the logit values of the generated sample $\hat{\textbf{x}}_i$. $CE(\cdot)$ denotes the cross-entropy function. The parameters $\textrm{W}_{fe}, \textrm{W}_{fc}$ of the global model are frozen. By minimizing $\mathcal{L}_{sem}$, the generator is forced to generate pseudo data to capture the input space (data distribution) of the global model. 

\textbf{Diversity Regularization} If we only employ the loss $\mathcal{L}_{sem}$, the generator probably undergoes the mode collapse problem and fails to achieve a good performance~\cite{bang2021mggan}. This problem is caused by shortcut learning of deep neural networks~\cite{geirhos2020shortcut}. For example, given two condition codes $\textbf{z}_i$ and $\textbf{z}_j$, with the same class label  $\hat{\textbf{y}}$, the synthetic samples  $\hat{\textbf{x}}_i = \mathcal{G}(\textbf{z}_i, \textrm{W}_\mathcal{G})$ and $\hat{\textbf{x}}_j = \mathcal{G}(\textbf{z}_j, \textrm{W}_\mathcal{G})$ may be collapsed into a point (namely, repeating samples), leading to low data diversity. To tackle this issue, we propose a diversity regularization loss to dynamically penalize the distance between $\hat{\textbf{x}}_i$ and $\hat{\textbf{x}}_j$:
\begin{equation}
\mathcal{L}_{div} = \frac{|\textbf{z}_i -\textbf{z}_j|}{|\mathcal{G}(\textbf{z}_i, \textrm{W}_\mathcal{G})-\mathcal{G}(\textbf{z}_j, \textrm{W}_\mathcal{G})|}
\label{eq7}
\end{equation}
where $|\cdot|$ denotes the $L_1$ norm distance. We minimize the loss $\mathcal{L}_{div}$ to provide a greater punishment on the distance between $\hat{\textbf{x}}_i$ and $\hat{\textbf{x}}_j$ when $\textbf{z}_i$ and $\textbf{z}_j$ are closer, thus encouraging the generator to create diverse images.

\textbf{Distribution Regularization} To improve the training stability of the generator, we follow the previous methods~\cite{yin2020dreaming,liu2021source} to use a distribution regularization loss to align
feature map statistics of the synthetic data at the Batch Normalization (BN) layer with their running counterparts:
\begin{equation}
\mathcal{L}_{dis} = \sum_l\left(\|\mu_l(\hat{\textbf{x}})-\mu_l\| + \|\sigma_l^2(\hat{\textbf{x}})-\sigma_l^2 \|\right),
\label{eq8}
\end{equation}
where $\mu_l(\hat{\textbf{x}})$ and $\sigma_l^2(\hat{\textbf{x}})$ are are the batch-wise mean and variance
estimates of feature maps corresponding to the $l$-th BN layer of the generator $\mathcal{G}(\cdot)$. $\mu_l$ and $\sigma_l^2$ are the mean and variance of the $l$-th BN layer of the global model. $\|\cdot\|$ denotes the $L_2$ norm distance.

\textbf{Explainable Conditions} A common type of the condition $\textbf{z}_i$ is random noise sampled from standard normal distribution $\mathcal{N}(\textbf{0},\textbf{\textit{I}})$ in previous methods~\cite{zhu2021data,yu2023data}. The generator is expected to learn the mapping between the noise space and the input space of the global model by minimizing $\mathcal{L}_{sem}$.
However, there are two drawbacks to this strategy: 1) randomly-sampled noises do not have any semantic information and  interpretability; 2) the correspondence between random noises $\textbf{z}$ and $\hat{\textbf{y}}$ are unclear and enables the generator to produce a lot of low-quality or repeating samples. Due to the drawbacks, it is difficult to learn a well-behaved 
generator to capture the global distribution of the global model. To solve this problem, we propose to build the condition pair $(\textbf{z}_i, \hat{\textbf{y}}_i)$ based on the concept embedding space. 
In particular, we first randomly sample the pseudo label $\hat{\textbf{y}}_i$ from a uniform distribution. Then the corresponding $\textbf{z}_i$ is probabilistic concept 
embedding sampled from the distribution $\mathcal{N}(\bm{\mu}_{\hat{\textbf{y}}_i}, \bm{\Sigma}_{\hat{\textbf{y}}_i})$ of the $\hat{\textbf{y}}_i$-th class built in Eq.~(\ref{eq2}),  and is fed into the generator:
\begin{equation}
\hat{\textbf{x}}_i = \mathcal{G}(\textbf{z}_i, \textrm{W}_\mathcal{G}), \ \  \textbf{z}_i \thicksim \mathcal{N}(\bm{\mu}_{\hat{\textbf{y}}_i}, \bm{\Sigma}_{\hat{\textbf{y}}_i}),
\label{eq9}
\end{equation}
where $\hat{\textbf{y}}_i$ is the label of the sample $\hat{\textbf{x}}_i$ and $\hat{\textbf{y}}_i \in [1, K]$. Compared with the distribution $\mathcal{N}(\textbf{0},\textbf{\textit{I}})$, the concept embedding distributions $\mathcal{N}(\bm{\mu}, \bm{\Sigma})$ are organized and class anchor-centered, making it easier for the generator to learn the mapping between the latent embedding space and the input space of global model. It is worth mentioning that the training of the generator only relies on the global model and concept embedding distributions, without requiring the private data of clients, thereby minimizing the communication and computational load.
\begin{algorithm}[!t]
    \label{algorithm1}
    \caption{The FedBM framework for heterogeneous FL.}
    \begin{algorithmic}[1]
        \renewcommand{\algorithmicrequire}{\textbf{Input:}}
        \renewcommand{\algorithmicensure}{\textbf{Output:}}
        \REQUIRE The client number $C$,  the number $E$ of local epochs, the local datasets $\{\mathcal{D}^c\}_{c=1}^C$.
        \STATE \textbf{Server executes:}
        \STATE //\textit{Constructing the global classifier via LKCC}
        \STATE Collecting concepts from clients to form a set $\{P_k\}_{k=1}^{K}$,
        \STATE Using PLMs to obtain a set of concept embeddings $\bm{E}$,
        \STATE Building concept embedding distributions via Eq.~(\ref{eq2}),
        \STATE Defining the global classifier via Eq.~(\ref{eq5}), 
        \STATE //\textit{Broadcasting the global classifier only once}
        \FOR{each communication round}
        \STATE //\textit{Estimating global distribution via CGDE}
        \STATE Training a generator $\mathcal{G}(\cdot)$ via Eq.~(\ref{eq10}), 
        \FOR{each client $c = 1,2,...,C$}
        \STATE $\textrm{W}_{fe}^c$ $\leftarrow$  ClientUpdate($\mathcal{G}, \textrm{W}_{fe}, \textrm{W}_{fc}$).
        \ENDFOR
        \STATE Model aggregation:  $\textrm{W}_{fe} \leftarrow \sum_{c=1}^C \textrm{W}_{fe}^c$.
        \ENDFOR
        \STATE \textbf{Client executes:}
        \STATE ClientUpdate($\mathcal{G}$, $\textrm{W}_{fe}$, $\textrm{W}_{fc}$):
        \STATE ~~~~Using $\textrm{W}_{fe}$ to initialize $\textrm{W}_{fe}^c$,
        \STATE ~~~~Using $\mathcal{G}$ to generate the synthesized dataset $\hat{\mathcal{D}}^c$, 
        \STATE ~~~~\textbf{for} {each epoch $e = 1,2,...,E$} \textbf{do}
        \STATE ~~~~~~~~$\min_{\textrm{W}_{fe}^c} \mathbb{E}_{(\textbf{x}_i, \textbf{y}_i)\thicksim \mathcal{D}^c \cup  \hat{\mathcal{D}}^c} \overline{\mathcal{L}}^\infty_{align}(\mathcal{F}(\textbf{x}_i, \textrm{W}_{fe}^c, \textrm{W}_{fc}), \textbf{y}_i)$.
        \STATE ~~~~\textbf{end for}
        \ENSURE The global model $\mathcal{F}(\textrm{W}_{fe}, \textrm{W}_{fc})$.
    \end{algorithmic}
    \label{algorithm1}
\end{algorithm}

\subsection{Optimization and Theoretical Derivation}
\label{method:optimization}
We present the pseudo code of the proposed FedBM framework in Algorithm.~\ref{algorithm1}. The training for FedBM includes the optimization of the generator and local models. The generator $\mathcal{G}$ is supervised by the following hybrid loss function, including the three regularization terms:
\begin{equation}
\mathcal{L}_{generator} = \mathcal{L}_{sem} +  \lambda_{div} \mathcal{L}_{div} + \lambda_{dis}  \mathcal{L}_{dis}.
\label{eq10}
\end{equation}
With these regularization terms, the generator is able to generate diverse data to capture the global data distribution of the global model. In experiments, we update the generator with multiple rounds of communication as a cycle to reduce computational and communication costs.
The updated generator $\mathcal{G}$ is sent to all clients to synthesize pseudo data. Assuming that $\hat{\mathcal{D}}^c$ is the local synthesized dataset at the $c$-th client and is updated in each round of training, we utilize $\hat{\mathcal{D}}^c$ and the local original data $\mathcal{D}^c$ to train the local feature extractor $\textrm{W}_{fe}^c$ by minimizing the surrogate loss $\overline{\mathcal{L}}^\infty_{align}$: 
\begin{equation}
\min_{\textrm{W}_{fe}^c} \mathbb{E}_{(\textbf{x}_i, \textbf{y}_i)\thicksim \mathcal{D}^c \cup  \hat{\mathcal{D}}^c} \overline{\mathcal{L}}^\infty_{align}(\mathcal{F}(\textbf{x}_i, \textrm{W}_{fe}^c, \textrm{W}_{fc}), \textbf{y}_i),
\label{eq11}
\end{equation}
where $\textbf{x}_i$ is sampled from the union of $\mathcal{D}^c$ and $\hat{\mathcal{D}}^c$. Since the generator contains global knowledge and is shared by all clients, it can guide the optimization of these clients toward a consistent direction. In addition, the generator is very lightweight and does not incur a big communication overhead. 

\section{Experiment}
\label{sec:experiments}
We present the experimental setup, the comparison results against previous methods, and the ablation results.
\subsection{Experiment Setup}
\subsubsection{Datasets}
To investigate the effectiveness of our FedBM framework, we evaluated it in five public datasets.

\textbf{OCT-C8}~\cite{subramanian2022classification} consists of 24,000 retinal OCT images and is divided into eight (categoriesage-related macular degeneration, choroidal neovascularisation, diabetic macular edema, drusen, macular hole, diabetic retinopathy, central serous retinopathy and one for healthy class). Based on the official division, 18,400 images are used for training, 2,800 for validation, and 2,800 for testing. 

\textbf{Kvasir-v2}~\cite{pogorelov2017kvasir} contains 8,000 endoscopic images of the gastrointestinal tract. These images belong to 8 categories, \textit{i.e.}, esophagitis, cecum, pylorus, Z-line, polyps, ulcerative colitis, dyed lifted polyp, dyed resection margin. We randomly partition these samples into training, validation, and test sets with a ratio of 7 : 1 : 2. 

\textbf{HAM1000}~\cite{tschandl2018ham10000} has 10,015 dermatoscopic images, which are from different populations. These images belong to 8 categories: actinic keratoses or intraepithelial carcinoma (akiec), basal cell carcinoma, benign keratosis-like lesions, dermatofibroma, melanoma, melanocytic nevi, and vascular lesions. Following the existing work~\cite{yang2023medmnist}, 7,007 images were used for training, 1,003 for validation, and 2,005 for testing. 

\textbf{PBC}~\cite{acevedo2020dataset} contains a total of 17,092 microscopic peripheral blood cell images. The dataset is organized into the eight groups: neutrophils, eosinophils, basophils, lymphocytes, monocytes, immature granulocytes, erythroblasts and platelets or thrombocytes. Following the existing work~\cite{yang2023medmnist}, 11,959 images are used for training, 1,712 for validation, and 3,421 for testing. 

\textbf{FEMNIST}~\cite{caldas2018leaf} is a part of the LEAF benchmark. It comprises 814,277 handwritten digit, lowercase, and uppercase letter images from 3,597 users, belonging to 62 classes. To ensure data heterogeneity, we select users whose sample sizes are fewer than 150, resulting in 396 users. For each experiment, 50 users are randomly selected from this group to conduct federated training. They  are treated as individual client and randomly divided into training set (35 users), validation set (5 users), and test set (10 users) with a ratio of 7: 1: 2. 

\subsubsection{Implementation Details}
The proposed FedBM and comparison methods are implemented with PyTorch library. We adopt the ResNet-18 \cite{he2016deep} as the backbone network of all methods.
The number of clients is set to 12 for OCT-C8 and PBC, and 10 for Kvasir-v2 and HAM1000 datasets, respectively.
The numbers of local epochs and communication rounds are set to 2 and 200 for OCT-C8 and Kvasir-v2 datasets, 1 and 200 for HAM1000 dataset, 10 and 50 for PBC dataset, and 5 and 200 for FEMNIST dataset, respectively. For all datasets, we utilize the Adam~\cite{kingma2014adam} optimizer with the initial learning rate of $1\times10^{-2}$. The batch size is set to 8 and the learning rate decays at a rate of 0.99 per epoch. 
The client sampling ratio is 0.5 except for FEMNIST (0.1).  Similar to existing FL works \cite{luo2021no,fedETF}, we use Dirichlet distribution on label ratios to simulate the Non-IID data distribution among clients. We set the Dirichlet parameter $\beta$ as 0.05 and 0.1 to ensure the high data heterogeneity. The weight $\lambda_{div}$ is set to 1.  $\lambda_{dis}$ and the batchsize of local synthesized datasets are selected from sets $[0.1, 1]$ and $[8, 16, 32]$ by grid search, respectively. The predetermined prompts can refer to the work~\cite{Zhu_Stealing_MICCAI2024}.
Two commonly-used metrics, accuracy, and F1 score, are used to measure the classification performance. In all the experiments, we conduct three trials for each setting and present the mean and the standard deviation. 

\begin{table*}[!t]
\caption{The performance comparison of the proposed method and existing methods on OCT-C8 dataset.}
\vspace{-2mm}
\center
\renewcommand\arraystretch{1}
\setlength{\tabcolsep}{4pt}
\begin{tabular}{p{80pt}|p{60pt}|p{60pt}|p{60pt}|p{60pt}}
\toprule[1pt]
 \multirow{2}{*}{Methods}& \multicolumn{2}{l|}{\makecell[c]{$\beta=0.05$}} & \multicolumn{2}{l}{\makecell[c]{$\beta=0.1$}} \\
 \cline{2-5}
 &    \makecell[c]{ Accuracy (\%) }     &   \makecell[c]{F1-score (\%)}       &    \makecell[c]{Accuracy (\%)}       &     \makecell[c]{F1-score (\%)}     \\
 \hline
 Central Learning &  \makecell[c]{$93.85\pm0.24$}    &     \makecell[c]{$93.84\pm0.22$}        &  \makecell[c]{$93.85\pm0.24$}  &    \makecell[c]{$93.84\pm0.22$}  \\
 \hline
FedAvg~\cite{fedavg}&    \makecell[c]{$74.82\pm5.98$}       &     \makecell[c]{$72.34\pm6.87$}        &      \makecell[c]{$78.64\pm5.44$}        &     \makecell[c]{$76.25\pm7.23$}        \\
FedDyn~\cite{acar2021federated}&     \makecell[c]{$70.79\pm1.94$}         &    \makecell[c]{$65.88\pm5.12$}         &     \makecell[c]{$73.46\pm5.49$}         &    \makecell[c]{$69.86\pm7.96$}         \\
FedProx~\cite{fedprox}&     \makecell[c]{$76.60\pm5.43$}         &    \makecell[c]{$74.49\pm6.12$}         &     \makecell[c]{$78.37\pm5.76$}         &    \makecell[c]{$75.64\pm7.68$}         \\
FedREP~\cite{collins2021exploiting}&     \makecell[c]{$43.87\pm7.24$}         &    \makecell[c]{$32.98\pm8.83$}         &     \makecell[c]{$59.37\pm12.81$}         &    \makecell[c]{$55.20\pm12.56$}         \\
FedROD~\cite{chen2021bridging}&     \makecell[c]{$70.20\pm4.17$}         &    \makecell[c]{$64.24\pm4.21$}         &     \makecell[c]{$79.11\pm5.62$}         &    \makecell[c]{$77.65\pm7.17$}         \\
FedNH~\cite{dai2023tackling}&     \makecell[c]{$80.08\pm4.23$}         &    \makecell[c]{$78.61\pm5.25$}         &      \makecell[c]{$82.53\pm2.57$}        &     \makecell[c]{$82.25\pm2.82$}        \\
FedProto~\cite{tan2022fedproto}&     \makecell[c]{$28.86\pm2.15$}         &    \makecell[c]{$16.95\pm1.79$}         &      \makecell[c]{$37.07\pm1.40$}        &     \makecell[c]{$24.88\pm2.32$}        \\
FedETF~\cite{fedETF}&     \makecell[c]{$77.79\pm5.17$}         &    \makecell[c]{$74.47\pm8.64$}         &      \makecell[c]{$82.81\pm3.76$}        &     \makecell[c]{$81.67\pm5.31$}        \\
Scaffold~\cite{karimireddy2020scaffold}&   \makecell[c]{$72.96\pm5.22$}  &  \makecell[c]{$70.55\pm7.80$} &  \makecell[c]{$79.34\pm3.97$}  &  \makecell[c]{$78.38\pm4.78$}  \\
 \hline
FedBM&     \makecell[c]{$\textbf{84.84}\pm2.77$}         &    \makecell[c]{$\textbf{84.55}\pm2.78$}         &     \makecell[c]{$\textbf{88.32}\pm1.49$}         &    \makecell[c]{$\textbf{88.16}\pm1.44$}         \\
\bottomrule[1pt]
\end{tabular}
\label{tab:OCT-C8}
\vspace{-3mm}
\end{table*}

\begin{table*}[!t]
\caption{The performance comparison of the proposed method and existing methods on Kvasir-v2 dataset.}
\vspace{-2mm}
\center
\renewcommand\arraystretch{1}
\setlength{\tabcolsep}{4pt}
\begin{tabular}{p{80pt}|p{60pt}|p{60pt}|p{60pt}|p{60pt}}
\toprule[1pt]
 \multirow{2}{*}{Methods}& \multicolumn{2}{l|}{\makecell[c]{$\beta=0.05$}} & \multicolumn{2}{l}{\makecell[c]{$\beta=0.1$}} \\
 \cline{2-5}
 &    \makecell[c]{ Accuracy (\%) }     &   \makecell[c]{F1-score (\%)}       &    \makecell[c]{Accuracy (\%)}       &     \makecell[c]{F1-score (\%)}     \\
 \hline
 Central Learning &    \makecell[c]{$77.15\pm0.03$}    &   \makecell[c]{$77.01\pm0.10$}      &    \makecell[c]{$77.15\pm0.03$}   &   \makecell[c]{$77.01\pm0.10$}    \\ 
 \hline
FedAvg~\cite{fedavg}&     \makecell[c]{$60.10\pm5.76$}         &    \makecell[c]{$54.07\pm9.58$}         &     \makecell[c]{$67.02\pm1.72$}         &    \makecell[c]{$63.93\pm2.49$}    \\
FedDyn~\cite{acar2021federated}&      \makecell[c]{$55.20\pm3.03$}     &    \makecell[c]{$49.84\pm3.86$}   &     \makecell[c]{$63.18\pm2.19$}   &    \makecell[c]{$60.98\pm1.41$}     \\
FedProx~\cite{fedprox}&      \makecell[c]{$59.68\pm2.02$}         &    \makecell[c]{$53.11\pm3.26$}         &     \makecell[c]{$68.77\pm1.45$}    &    \makecell[c]{$66.81\pm2.53$}    \\
FedREP~\cite{collins2021exploiting}&     \makecell[c]{$33.06\pm15.71$}     &    \makecell[c]{$23.88\pm13.72$}   &     \makecell[c]{$48.95\pm0.62$}   &    \makecell[c]{$39.71\pm2.50$}    \\
FedROD~\cite{chen2021bridging}&      \makecell[c]{ $61.79\pm2.72$}   &    \makecell[c]{$58.83\pm3.40$}    &     \makecell[c]{$70.10\pm3.70$}   &    \makecell[c]{$68.01\pm5.71$}    \\
FedNH~\cite{dai2023tackling}&     \makecell[c]{$61.08\pm5.68$}         &    \makecell[c]{$54.15\pm8.68$}         &      \makecell[c]{$69.10\pm1.74$}        &     \makecell[c]{$65.71\pm3.32$}        \\
FedProto~\cite{tan2022fedproto}&     \makecell[c]{$26.79\pm1.97$}         &    \makecell[c]{$14.99\pm1.37$}         &      \makecell[c]{$35.32\pm1.59$}        &     \makecell[c]{$23.33\pm1.30$}        \\
FedETF~\cite{fedETF}&     \makecell[c]{$63.77\pm3.73$}         &    \makecell[c]{$60.12\pm6.34$}         &     \makecell[c]{$69.70\pm3.56$}         &    \makecell[c]{$67.15\pm6.64$}   \\
Scaffold~\cite{karimireddy2020scaffold}&   \makecell[c]{$59.06\pm5.41$}  &  \makecell[c]{$54.25\pm5.41$} &  \makecell[c]{$66.16\pm1.80$}  &  \makecell[c]{$64.88\pm1.46$}  \\
 \hline
FedBM&     \makecell[c]{$\textbf{71.31}\pm2.81$}         &    \makecell[c]{$\textbf{69.51}\pm4.11$}         &     \makecell[c]{$\textbf{74.14}\pm 1.89$}         &    \makecell[c]{$\textbf{73.75}\pm1.79$}         \\
\bottomrule[1pt]
\end{tabular}
\label{tab:Kvasir-v2}
\vspace{-3mm}
\end{table*}

\begin{table*}[!t]
\caption{The performance comparison of the proposed method and existing methods on HAM1000 dataset.}
\vspace{-2mm}
\center
\renewcommand\arraystretch{1}
\setlength{\tabcolsep}{4pt}
\begin{tabular}{p{80pt}|p{60pt}|p{60pt}|p{60pt}|p{60pt}}
\toprule[1pt]
 \multirow{2}{*}{Methods}& \multicolumn{2}{l|}{\makecell[c]{$\beta=0.05$}} & \multicolumn{2}{l}{\makecell[c]{$\beta=0.1$}} \\
 \cline{2-5}
 &    \makecell[c]{ Accuracy (\%) }     &   \makecell[c]{F1-score (\%)}       &    \makecell[c]{Accuracy (\%)}       &     \makecell[c]{F1-score (\%)}     \\
 \hline
Central Learning &    \makecell[c]{$74.05\pm2.19$}       &     \makecell[c]{$51.03\pm0.91$}        &  \makecell[c]{$74.05\pm2.19$}     &    \makecell[c]{$51.03\pm0.91$}  \\
 \hline 
FedAvg~\cite{fedavg}&     \makecell[c]{$67.43\pm3.59$}         &    \makecell[c]{$31.84\pm1.90$}         &      \makecell[c]{$68.77\pm3.32$}        &     \makecell[c]{$41.25\pm1.86$}        \\
FedDyn~\cite{acar2021federated}&      \makecell[c]{$67.84\pm1.14$}         &    \makecell[c]{$20.33\pm0.70$}         &      \makecell[c]{$66.48\pm4.01$}        &     \makecell[c]{$35.79\pm 1.14$}        \\
FedProx~\cite{fedprox}&      \makecell[c]{$67.56\pm2.64$}         &    \makecell[c]{$35.53\pm6.32$}         &      \makecell[c]{$69.42\pm0.77$}        &     \makecell[c]{$38.03\pm7.17$}        \\
FedREP~\cite{collins2021exploiting}&    \makecell[c]{$61.74\pm2.57$}         &    \makecell[c]{$25.67\pm1.99$}         &      \makecell[c]{$59.98\pm4.04$}        &     \makecell[c]{$30.57\pm2.04$}        \\
FedROD~\cite{chen2021bridging}&     \makecell[c]{$60.89\pm1.03$}         &    \makecell[c]{$37.90\pm4.77$}         &      \makecell[c]{$67.99\pm0.61$}        &     \makecell[c]{$46.08\pm0.87$}        \\
FedNH~\cite{dai2023tackling}&     \makecell[c]{$64.95\pm4.62$}         &    \makecell[c]{$26.45\pm5.48$}         &      \makecell[c]{$58.28\pm 9.37$}        &     \makecell[c]{$35.97\pm5.00$}        \\
FedProto~\cite{tan2022fedproto}&     \makecell[c]{$36.62\pm2.53$}         &    \makecell[c]{$11.50\pm1.15$}         &      \makecell[c]{$41.06\pm1.12$}        &     \makecell[c]{$13.72\pm1.28$}        \\
FedETF~\cite{fedETF}&    \makecell[c]{$67.36\pm1.16$}         &    \makecell[c]{$33.67\pm4.33$}         &      \makecell[c]{$64.67\pm2.26$}        &     \makecell[c]{$43.45\pm0.84$}        \\
Scaffold~\cite{karimireddy2020scaffold}&   \makecell[c]{$67.88\pm1.36$}  &  \makecell[c]{$27.19\pm5.05$} &  \makecell[c]{$66.96\pm2.83$}  &  \makecell[c]{$36.99\pm4.35$}  \\
 \hline
FedBM&    \makecell[c]{$\textbf{72.75}\pm0.88$}         &    \makecell[c]{$\textbf{45.43}\pm3.13$}         &      \makecell[c]{$\textbf{73.55}\pm0.28$}        &     \makecell[c]{$\textbf{49.49}\pm1.74$}        \\ 
\bottomrule[1pt]
\end{tabular}
\label{tab:HAM1000}
\vspace{-3mm}
\end{table*}

\subsection{Comparison with State-of-the-art Methods}
We compare our FedBM framework with the state-of-the-art FL approaches in four datasets, including FedAvg~\cite{fedavg}, FedDyn~\cite{acar2021federated}, FedProx~\cite{fedprox}, FedREP~\cite{collins2021exploiting},  FedROD~\cite{chen2021bridging}, FedNH~\cite{dai2023tackling}, FedProto~\cite{tan2022fedproto}, FedETF~\cite{fedETF} and Scaffold~\cite{karimireddy2020scaffold}. For a fair comparison, these methods are implemented in the standard FL framework, using the same split as FedBM for each dataset. The common hyperparameters for all methods are consistent and are determined by FedAvg~\cite{fedavg}, such as learning rate, number of training round and so on. Following FedETF, we set $\mu$ to 0.001 in FedPROX and 0.01 in FedDyn as suggested in their official implementations, and set $\gamma$ = 1 in FedROD. For FedNH, the smoothing hyperparameter $\rho$ = 0.9 as suggested in the original paper.

\begin{table*}[!t]
\caption{The performance comparison of the proposed method and existing methods on PBC dataset.}
\vspace{-2mm}
\center
\renewcommand\arraystretch{1}
\setlength{\tabcolsep}{4pt}
\begin{tabular}{p{80pt}|p{60pt}|p{60pt}|p{60pt}|p{60pt}}
\toprule[1pt]
 \multirow{2}{*}{Methods}& \multicolumn{2}{l|}{\makecell[c]{$\beta=0.05$}} & \multicolumn{2}{l}{\makecell[c]{$\beta=0.1$}} \\
 \cline{2-5}
 &    \makecell[c]{ Accuracy (\%) }     &   \makecell[c]{F1-score (\%)}       &    \makecell[c]{Accuracy (\%)}       &     \makecell[c]{F1-score (\%)}     \\
 \hline
Central Learning &    \makecell[c]{$97.43\pm0.06$}       &     \makecell[c]{$97.26\pm0.09$}        &   \makecell[c]{$97.43\pm0.06$}    & \makecell[c]{$97.26\pm0.09$}     \\
 \hline 
FedAvg~\cite{fedavg}&     \makecell[c]{$63.11\pm8.13$}         & \makecell[c]{$47.98\pm9.37$}     &\makecell[c]{$78.61\pm12.65$}   & \makecell[c]{$71.73\pm16.50$}     \\
FedDyn~\cite{acar2021federated}&   \makecell[c]{$21.18\pm3.89$}         & \makecell[c]{$6.19\pm2.83$}     &\makecell[c]{$28.90\pm6.28$}   & \makecell[c]{$15.76\pm5.04$}     \\
FedProx~\cite{fedprox}&  \makecell[c]{$79.51\pm0.96$}         & \makecell[c]{$72.49\pm3.33$}     &\makecell[c]{$87.37\pm5.34$}   & \makecell[c]{$85.39\pm7.05$}     \\
FedREP~\cite{collins2021exploiting}&   \makecell[c]{$37.49\pm2.61$}         & \makecell[c]{$26.47\pm3.18$}     &\makecell[c]{$42.74\pm7.21$}   & \makecell[c]{$33.66\pm9.70$}     \\
FedROD~\cite{chen2021bridging}&  \makecell[c]{$75.89\pm1.61$}         & \makecell[c]{$67.89\pm2.03$}     &\makecell[c]{$79.81\pm12.38$}   & \makecell[c]{$77.87\pm13.18$}     \\
FedNH~\cite{dai2023tackling}&  \makecell[c]{$74.36\pm5.16$}         & \makecell[c]{$64.22\pm5.63$}     &\makecell[c]{$84.69\pm7.32$}   & \makecell[c]{$82.44\pm8.47$}     \\
FedProto~\cite{tan2022fedproto}&  \makecell[c]{$30.24\pm2.23$}         & \makecell[c]{$18.17\pm1.70$}     &\makecell[c]{$41.15\pm2.52$}   & \makecell[c]{$28.28\pm1.88$}     \\
FedETF~\cite{fedETF}&  \makecell[c]{$77.91\pm2.23$}         & \makecell[c]{$69.78\pm4.90$}     &\makecell[c]{$87.86\pm4.19$}   & \makecell[c]{$84.33\pm6.73$}     \\
Scaffold~\cite{karimireddy2020scaffold}&   \makecell[c]{$58.03\pm7.06$}  &  \makecell[c]{$45.55\pm4.84$} &  \makecell[c]{$\textbf{93.65}\pm2.42$}  &  \makecell[c]{$\textbf{93.35}\pm2.52$}  \\
 \hline
FedBM&   \makecell[c]{$\textbf{84.90}\pm3.29$}         & \makecell[c]{$\textbf{80.54}\pm5.53$}     &\makecell[c]{${89.88}\pm4.47$}   & \makecell[c]{${87.71}\pm6.09$}     \\
\bottomrule[1pt]
\end{tabular}
\label{tab:PBC}
\vspace{-3mm}
\end{table*}

\begin{table}[h]
\caption{The performance comparison of the proposed method and existing methods on FEMNIST dataset.}
\center
\renewcommand\arraystretch{1}
\setlength{\tabcolsep}{1pt}
\begin{tabular}{p{80pt}|p{56pt}|p{56pt}}
\toprule[1pt]
Methods & \makecell[c]{ Accuracy (\%) }  & \makecell[c]{F1-score (\%)}  \\
 \hline
Central Learning &  \makecell[c]{$79.13\pm1.27$}  &  \makecell[c]{$58.89\pm1.35$}   \\
 \hline
FedAvg~\cite{fedavg}&   \makecell[c]{$73.06\pm2.09$}  &  \makecell[c]{$53.83\pm3.58$}   \\
FedDyn~\cite{acar2021federated}&   \makecell[c]{$69.17\pm2.41$}  &  \makecell[c]{$44.04\pm2.44$}  \\
FedProx~\cite{fedprox}&   \makecell[c]{$73.60\pm1.70$}  &  \makecell[c]{$50.99\pm1.65$}  \\
FedREP~\cite{collins2021exploiting}&    \makecell[c]{$50.92\pm2.58$}  &  \makecell[c]{$22.71\pm1.74$}  \\
FedROD~\cite{chen2021bridging}&   \makecell[c]{$71.66\pm0.53$}  &  \makecell[c]{$51.48\pm2.35$} \\
FedNH~\cite{dai2023tackling}&   \makecell[c]{$74.42\pm2.02$}  &  \makecell[c]{$52.34\pm2.01$}  \\
FedProto~\cite{tan2022fedproto}&   \makecell[c]{$28.57\pm1.16$}  &  \makecell[c]{$10.19\pm0.22$}  \\
FedETF~\cite{fedETF}&   \makecell[c]{$73.48\pm1.17$}  &  \makecell[c]{$53.98\pm1.13$}   \\
Scaffold\cite{karimireddy2020scaffold}&   \makecell[c]{$74.55\pm1.19$}  &  \makecell[c]{$51.61\pm1.37$}   \\
 \hline
FedBM&   \makecell[c]{$\textbf{75.98}\pm2.15$}  &  \makecell[c]{$\textbf{54.77}\pm0.81$}   \\
\bottomrule[1pt]
\end{tabular}
\label{tab:FEMNIST}
\end{table}

\begin{table*}[!t]
\caption{The performance of the proposed FedBM framework with different modules. 
}
\center
\renewcommand\arraystretch{1}
\setlength{\tabcolsep}{4pt}
\begin{tabular}{p{40pt}|p{60pt}|p{60pt}|p{60pt}|p{60pt}|p{60pt}}
\toprule
\multirow{2}{1.1cm}{\makecell[c]{Datasets} }&\multirow{2}{2cm}{\makecell[c]{Methods} }      & \multicolumn{2}{l|}{\makecell[c]{$\beta=0.05$ }} & \multicolumn{2}{l}{\makecell[c]{$\beta=0.1$}} \\
\cmidrule(r){3-6}
&                 &    \makecell[c]{Accuracy (\%)}     &   \makecell[c]{F1-score (\%)}   &   \makecell[c]{Accuracy (\%)}        &  \makecell[c]{F1-score (\%)}  \\
\midrule
\multirow{4}{0cm}{\makecell[c]{OCT-8} }&\makecell[c]{w/o LKCC }&  \makecell[c]{$83.02\pm2.45$}  & \makecell[c]{$82.94\pm2.47$} & \makecell[c]{$85.20\pm3.54$} & \makecell[c]{$84.92\pm3.72$}  \\
&\makecell[c]{w/o CGDE} &   \makecell[c]{\textbf{$79.14\pm3.77$}}         &    \makecell[c]{\textbf{$77.13\pm5.10$}}         &     \makecell[c]{\textbf{$85.00\pm2.66$}}         &    \makecell[c]{\textbf{$84.56\pm2.99$}}         \\
&\makecell[c]{FedBM}& \makecell[c]{\textbf{$84.84\pm2.77$}}         &    \makecell[c]{\textbf{$84.55\pm2.78$}}         &     \makecell[c]{\textbf{$88.32\pm1.49$}}         &    \makecell[c]{\textbf{$88.16\pm1.44$}}         \\
\midrule
\multirow{4}{0cm}{\makecell[c]{Kvasir-v2} }&\makecell[c]{w/o LKCC }&  \makecell[c]{$28.04\pm9.74$}  & \makecell[c]{$21.18\pm7.12$} & \makecell[c]{$23.68\pm5.79$} & \makecell[c]{$16.07\pm7.13$}  \\
&\makecell[c]{w/o CGDE} &  \makecell[c]{\textbf{$65.00\pm4.36$}}    &    \makecell[c]{\textbf{$61.67\pm6.61$}}   &     \makecell[c]{\textbf{$70.90\pm1.97$}}         &    \makecell[c]{\textbf{$68.73\pm3.36$}}   \\ 
&\makecell[c]{FedBM}&  \makecell[c]{\textbf{$71.31\pm2.81$}}         &    \makecell[c]{\textbf{$69.51\pm4.11$}}         &     \makecell[c]{\textbf{$74.14\pm 1.89$}}         &    \makecell[c]{\textbf{$73.75\pm1.79$}}         \\
 \bottomrule
\end{tabular}
\label{tab:ablation_submodule}
\end{table*}

\begin{table*}[!t]
\caption{The performance of LKCC with different proportions of prompts. LKCC (25\%) indicates that only 25\% of prompts are used.}
\center
\renewcommand\arraystretch{1}
\setlength{\tabcolsep}{6pt}
\begin{tabular}{p{65pt}|p{60pt}|p{60pt}|p{60pt}|p{60pt}}
\toprule[1pt]
 \multirow{2}{*}{Methods}& \multicolumn{2}{l|}{\makecell[c]{$\beta=0.05$}} & \multicolumn{2}{l}{\makecell[c]{$\beta=0.1$}} \\
 \cline{2-5}
 &    \makecell[c]{ Accuracy (\%) }     &   \makecell[c]{F1-score (\%)}       &    \makecell[c]{Accuracy (\%)}       &     \makecell[c]{F1-score (\%)}     \\
 \hline
LKCC (25\%)&   \makecell[c]{$76.64\pm6.26$}       &     \makecell[c]{$76.05\pm6.48$}        &      \makecell[c]{$83.71\pm2.83$}        &     \makecell[c]{$83.45\pm3.05$}        \\
LKCC (50\%)&      \makecell[c]{$76.82\pm5.41$}       &     \makecell[c]{$76.54\pm4.88$}        &      \makecell[c]{$82.40\pm3.17$}        &     \makecell[c]{$82.23\pm3.30$}        \\
LKCC (75\%)&      \makecell[c]{$77.70\pm3.53$}       &     \makecell[c]{$75.71\pm4.51$}        &      \makecell[c]{$84.12\pm4.18$}        &     \makecell[c]{$83.39\pm4.79$}        \\
LKCC (100\%)&      \makecell[c]{$79.14\pm3.77$}         &    \makecell[c]{$77.13\pm5.10$}         &     \makecell[c]{$85.00\pm2.66$}         &    \makecell[c]{$84.56\pm2.99$}         \\
\bottomrule[1pt]
\end{tabular}
\label{tab:ablation_prompt_number}
\end{table*}

\begin{table*}[!t]
\caption{The performance of different methods on OCT-C8 dataset.}
\center
\renewcommand\arraystretch{1}
\setlength{\tabcolsep}{4pt}
\begin{tabular}{p{100pt}|p{60pt}|p{60pt}|p{60pt}|p{60pt}}
\toprule[1pt]
 \multirow{2}{*}{Methods}& \multicolumn{2}{l|}{\makecell[c]{$\beta=0.05$}} & \multicolumn{2}{l}{\makecell[c]{$\beta=0.1$}} \\
 \cline{2-5}
 &    \makecell[c]{ Accuracy (\%) }     &   \makecell[c]{F1-score (\%)}       &    \makecell[c]{Accuracy (\%)}       &     \makecell[c]{F1-score (\%)}     \\
 \hline
Baseline&   \makecell[c]{$74.82\pm5.98$}       &     \makecell[c]{$72.34\pm6.87$}        &      \makecell[c]{$78.64\pm5.44$}        &     \makecell[c]{$76.25\pm7.23$}        \\
Random (Freezing)&    \makecell[c]{$76.11\pm6.31$}         &    \makecell[c]{$73.07\pm7.57$}         &     \makecell[c]{$82.67\pm3.40$}         &    \makecell[c]{$82.03\pm3.80$}         \\
Embedding (Averaging) &      \makecell[c]{$76.75\pm5.00$}         &    \makecell[c]{$76.34\pm4.38$}         &     \makecell[c]{$84.22\pm2.38$}         &    \makecell[c]{$83.81\pm2.64$}         \\
Embedding (Distribution) &      \makecell[c]{$79.14\pm3.77$}         &    \makecell[c]{$77.13\pm5.10$}         &     \makecell[c]{$85.00\pm2.66$}         &    \makecell[c]{$84.56\pm2.99$}         \\
\bottomrule[1pt]
\end{tabular}
\label{tab:ablation_prompt_method}
\end{table*}

\textbf{Comparison Results on OCT-C8:} In Table~\ref{tab:OCT-C8}, we show the classification performance of different methods on OCT-C8 dataset to validate the proposed FedBM. Although both FedProx and FedDyn introduce constraints to the parameters of local models, the latter achieves better performance. Our FedBM suppresses FedProx with significant performance increments for two cases, such as $9.65\%$ in F1-score for $\beta=0.05$ and $12.52\%$ in Accuracy for $\beta=0.1$. Furthermore, in contrast to FedETF, which employs orthogonal initialization to build local classifiers, our method obtains superior performance with a remarkable increments of 2.66\% ($\beta=0.05$) and 2.89\% ($\beta=0.1$) in F1-score. Noticeably, FedBM outperforms the second-best method, i.e., FedNH, by tremendous performance gaps for two cases, including $5.39\%$ in F1-score ($\beta=0.05$) and $5.79\%$ ($P$-value $ < 0.05$) in Accuracy ($\beta=0.1$). 

\textbf{Comparison Results on Kvasir-v2:} The performance of previous methods and FedBM on Kvasir-v2 dataset are demonstrated in Table~\ref{tab:Kvasir-v2}. It can be observed that all previous methods implement poor performance. FedREP and FedProto even fail to converge when the data of clients are seriously heterogeneous ($\beta=0.05$).  The proposed FedBM exceeds the second-best approach, FedETF, with the overwhelming performance advantages in two cases, such as $7.54\%$ ($P$-value $ < 0.06$) and $9.39$ ($P$-value $ < 0.01$), $4.44\%$ and $6.60\%$ in Accuracy and F1-score for $\beta=0.05$ and $\beta=0.1$, respectively. Meanwhile, FedETF also shows a larger performance drop ($5.93\%$ in Accuracy and $7.03\%$ in F1-score) from $\beta=0.1$ to $0.05$, while FedBM only undergoes $2.83\%$ in Accuracy and $4.24\%$ in F1-score. 

\textbf{Comparison Results on HAM1000:} Table~\ref{tab:HAM1000} presents the performance of existing methods and FedBM on HAM1000 dataset. Among existing methods, FedROD obtains a relatively good performance with $67.99\%$ in Accuracy and $46.08\%$ in F1-score when $\beta$ is $0.1$. However, when data becomes more heterogeneous ($\beta=0.05$), FedROD suffers from a significant performance degradation, merely achieving $60.89\%$ in Accuracy and $37.90\%$ in F1-score, with $7.10\%$ and $8.18\%$ of drops, respectively. By comparison, FedBM outperforms FedROD with 
remarkable performance improvements in two cases, such as $5.56\%$ and $11.86\%$ in Accuracy, $3.41\%$  and $7.53\%$ in F1-score for $\beta=0.05$ and $\beta=0.1$ ($P$-value $ < 0.05$), respectively. It is worth noting that FedBM only experiences a slight performance drops ($0.79\%$ in Accuracy and $4.06\%$ in F1-score).

\textbf{Comparison Results on PBC:} Table~\ref{tab:PBC} shows the performance of existing approaches and FedBM on PCB dataset. We can observe that FedDyn, FedREP and FedProto are difficult to converge, yielding very low performance. FedProx and FedETF achieve good performance ($87.37\%$ and $87.86\%$  in Accuracy, $85.39\%$ and $84.33\%$ in F1-score, respectively) when the heterogeneity parameter $\beta$ is $0.1$. They only show a small performance gap compared with FedBM ($89.88\%$ and $87.71\%$ in Accuracy and F1-score). FedBM remarkably surpasses FedProx and FedETF when $\beta$ is $0.05$, with prominent increments of $8.05\%$ and $6.99\%$ in Accuracy, $5.39\%$ and $10.76\%$ in F1-score, respectively. Although Scaffold exceeds FedBM when $\beta=0.1$, it undergoes a severe performance decline when $\beta$ becomes $0.05$, showing the vulnerability to data heterogeneity.

\textbf{Comparison Results on FEMNIST:} Table~\ref{tab:FEMNIST} presents the classification performance of different approaches on FEMNIST. It can be observed that FedProto obtains the worst performance, since only transmitting prototypes cannot handle the data heterogeneity problem when each client has limited data. FedNH performs better than FedProto because it not only transmits prototypes to calibrate classifiers but also aggregates feature extractors across clients. In comparison, the proposed FedBM outperforms FedNH by a considerable margin in both Accuracy ($1.56\%$) and F1-score ($2.43\%$). Among all methods, Scaffold yields the second-best accuracy score ($75.98\%$) but obtains the inferior result in F1-score. FedBM shows the best performance in both Accuracy ($75.98\%$) and F1-score ($54.77\%$).

These experimental results on the five datasets demonstrate the remarkable performance advantage of our method over state-of-the-art FL methods under different heterogeneous settings. Additionally, FedBM also shows more stable performance against the data heterogeneity than existing approaches.

\begin{table*}[!t]
\caption{The performance of the proposed FedBM framework with different modules. 
}
\center
\renewcommand\arraystretch{1}
\setlength{\tabcolsep}{4pt}

\begin{tabular}{p{40pt}|p{90pt}|p{60pt}|p{60pt}|p{60pt}|p{60pt}}
\toprule
\multirow{2}{1.1cm}{\makecell[c]{Datasets} }&\multirow{2}{3cm}{\makecell[c]{Methods}}       & \multicolumn{2}{l|}{\makecell[c]{$\beta=0.05$ }} & \multicolumn{2}{l}{\makecell[c]{$\beta=0.1$}} \\
\cmidrule(r){3-6}
&                 &    \makecell[c]{Accuracy (\%)}     &   \makecell[c]{F1-score (\%)}   &   \makecell[c]{Accuracy (\%)}        &  \makecell[c]{F1-score (\%)}  \\
\midrule
\multirow{4}{0cm}{\makecell[c]{OCT-8} }&\makecell[c]{w/o Concept Conditions} & \makecell[c]{$81.19\pm3.25$}  & \makecell[c]{$80.18\pm2.85$} & \makecell[c]{$84.09\pm4.28$} & \makecell[c]{$83.43\pm4.99$}  \\
&\makecell[c]{w/o $\mathcal{L}_{div}$} & \makecell[c]{$84.26\pm2.36$}  & \makecell[c]{$83.65\pm2.58$} & \makecell[c]{$87.60\pm1.88$} & \makecell[c]{$87.65\pm1.90$}  \\
&\makecell[c]{w/o $\mathcal{L}_{dis}$}&  \makecell[c]{$82.13\pm2.30$}  & \makecell[c]{$81.54\pm2.29$} & \makecell[c]{$85.91\pm1.67$} & \makecell[c]{$85.99\pm1.65$}  \\
&\makecell[c]{FedBM}& \makecell[c]{\textbf{$84.84\pm2.77$}}         &    \makecell[c]{\textbf{$84.55\pm2.78$}}         &     \makecell[c]{\textbf{$88.32\pm1.49$}}         &    \makecell[c]{\textbf{$88.16\pm1.44$}}         \\
\midrule
\multirow{4}{0cm}{\makecell[c]{Kvasir-v2} }&\makecell[c]{w/o Concept Conditions} & \makecell[c]{$64.29\pm1.56$}  & \makecell[c]{$63.05\pm1.87$} & \makecell[c]{$71.16\pm2.50$} & \makecell[c]{$70.25\pm3.12$}  \\
&\makecell[c]{w/o $\mathcal{L}_{div}$} & \makecell[c]{$70.00\pm3.12$}  & \makecell[c]{$68.14\pm4.10$} & \makecell[c]{$74.52\pm2.13$} & \makecell[c]{$74.03\pm2.32$}  \\
&\makecell[c]{w/o $\mathcal{L}_{dis}$}&\makecell[c]{$67.95\pm2.10$}  & \makecell[c]{$65.39\pm2.91$} & \makecell[c]{$72.10\pm0.78$} & \makecell[c]{$71.54\pm0.83$}  \\
&\makecell[c]{FedBM}&  \makecell[c]{\textbf{$71.31\pm2.81$}}         &    \makecell[c]{\textbf{$69.51\pm4.11$}}         &     \makecell[c]{\textbf{$74.14\pm 1.89$}}         &    \makecell[c]{\textbf{$73.75\pm1.79$}}         \\
 \bottomrule
\end{tabular}
\label{tab:ablation_CGDE}
\end{table*}

\subsection{Ablation Study}
\subsubsection{Evaluation of Different Modules}
LKCC and CGDE are two indispensable components of our FedBM framework to alleviate the data heterogeneity issue. To evaluate their contributions, we individually remove them to observe the performance of FedBM. As illustrated in Table~\ref{tab:ablation_submodule}, FedBM experiences significant performance decline once we remove LKCC (w/o LKCC), with decrements of $1.61\%$ ($\beta=0.05$) and $3.24\%$ ($\beta=0.1$) on OCT-8, and $48.33\%$ ($\beta=0.05$) and $57.68\%$ ($\beta=0.1$) on Kvasir-v2 in F1-score. Global model with a biased classifier leads to low-quality generated samples. Exploiting these samples to train local models incurs performance degradation.  
Moreover, we can observe that discarding CGDE (w/o CGDE) also leads to a substantial performance drop, with decrements of $5.70\%$ ($\beta=0.05$) and $3.32\%$ ($\beta=0.1$) on OCT-8, and $6.31\%$ ($\beta=0.05$) and $3.24\%$ ($\beta=0.1$) on Kvasir-v2 in Accuracy. The experimental results highlight the importance of aligning local feature extractors. The best results are obtained when FedBM is equipped with LKCC and CGDE, which can corroborate the effectiveness of the two modules.

\begin{table*}[!t]
\caption{The performance of the proposed FedBM with different $\lambda_{div}$.}
\center
\renewcommand\arraystretch{1.1}
\setlength{\tabcolsep}{4pt}
\begin{tabular}{p{40pt}|p{60pt}|p{60pt}|p{60pt}|p{60pt}|p{60pt}}
\toprule[1pt]
 \multirow{2}{*}{Datasets}& \multirow{2}{*}{\ \ \ \ \ \ \ \ \ $\lambda_{div}$}& \multicolumn{2}{l|}{\makecell[c]{$\beta=0.05$}} & \multicolumn{2}{l}{\makecell[c]{$\beta=0.1$}} \\
 \cline{3-6}
 &  &  \makecell[c]{ Accuracy }     &   \makecell[c]{F1-score}       &    \makecell[c]{Accuracy}       &     \makecell[c]{F1-score}     \\
 \hline
\multirow{3}{*}{OCT-C8} &  \makecell[c]{0.1} & \makecell[c]{$83.90\pm2.71$}& \makecell[c]{$83.64\pm2.71$} &\makecell[c]{$86.92\pm1.82$}& \makecell[c]{$87.10\pm1.68$} \\
\cline{2-6}
                  &  \makecell[c]{1.0}& \makecell[c]{$\textbf{84.84}\pm2.77$}         &    \makecell[c]{$\textbf{84.55}\pm2.78$}         &     \makecell[c]{$\textbf{88.32}\pm1.49$}         &    \makecell[c]{$\textbf{88.16}\pm1.44$}        \\
\cline{2-6}
                  &  \makecell[c]{10}& \makecell[c]{$83.26\pm3.02$} & \makecell[c]{$82.68\pm3.02$} & \makecell[c]{$87.51\pm1.94$} & \makecell[c]{$87.39\pm2.03$}\\                  
 \hline
\multirow{3}{*}{Kvasir-v2} &  \makecell[c]{0.1} & \makecell[c]{$69.31\pm2.96$}& \makecell[c]{$67.40\pm3.89$} &\makecell[c]{$73.77\pm1.13$}& \makecell[c]{$73.26\pm1.06$} \\
\cline{2-6}
                  &  \makecell[c]{1.0}& \makecell[c]{$\textbf{71.31}\pm2.81$}       &    \makecell[c]{$\textbf{69.51}\pm4.11$}         &     \makecell[c]{$74.14\pm 1.89$}        &    \makecell[c]{$\textbf{73.75}\pm1.79$}     \\
\cline{2-6}
                  &  \makecell[c]{10}& \makecell[c]{$69.62\pm2.35$} & \makecell[c]{$67.90\pm3.86$} & \makecell[c]{$\textbf{74.35}\pm1.69$} & \makecell[c]{$73.71\pm2.07$}\\
\bottomrule[1pt]
\end{tabular}
\label{tab:div}
\end{table*}

\begin{table*}[!t]
\caption{The performance of the proposed FedBM with different $\lambda_{dis}$.}
\center
\renewcommand\arraystretch{1.1}
\setlength{\tabcolsep}{4pt}
\begin{tabular}{p{60pt}|p{60pt}|p{60pt}|p{60pt}|p{60pt}|p{60pt}}
\toprule[1pt]
 \multirow{2}{*}{Datasets}& \multirow{2}{*}{\ \ \ \ \ \ \ \ \ $\lambda_{dis}$}& \multicolumn{2}{l|}{\makecell[c]{$\beta=0.05$}} & \multicolumn{2}{l}{\makecell[c]{$\beta=0.1$}} \\
 \cline{3-6}
 &  &  \makecell[c]{ Accuracy }     &   \makecell[c]{F1-score}       &    \makecell[c]{Accuracy}       &     \makecell[c]{F1-score}     \\
 \hline
\multirow{2}{*}{OCT-C8} &  \makecell[c]{0.1} & \makecell[c]{$\textbf{84.84}\pm2.77$}         &    \makecell[c]{$\textbf{84.55}\pm2.78$}     &\makecell[c]{$\textbf{88.32}\pm1.49$}        &    \makecell[c]{$\textbf{88.16}\pm1.44$}       \\
\cline{2-6}
                  &  \makecell[c]{1}& \makecell[c]{$83.50\pm4.00$} & \makecell[c]{$83.33\pm3.93$} & \makecell[c]{$87.05\pm3.12$} & \makecell[c]{$87.12\pm3.18$}\\   
\cline{2-6}
                  &  \makecell[c]{10}& \makecell[c]{$83.00\pm2.60$} & \makecell[c]{$82.46\pm2.75$} & \makecell[c]{$86.17\pm3.58$} & \makecell[c]{$86.33\pm3.47$}\\                    
 \hline
\multirow{2}{*}{Kvasir-v2} &  \makecell[c]{0.1} & \makecell[c]{$70.14\pm2.76$}& \makecell[c]{$67.72\pm4.34$} &\makecell[c]{$\textbf{74.14}\pm 1.89$}         &    \makecell[c]{$\textbf{73.75}\pm1.79$}    \\
\cline{2-6}
                  &  \makecell[c]{1}& \makecell[c]{$\textbf{71.31}\pm2.81$}        &    \makecell[c]{$\textbf{69.51}\pm4.11$}  & \makecell[c]{$72.33\pm1.81$} & \makecell[c]{$71.50\pm2.04$}\\
\cline{2-6}
                  &  \makecell[c]{10}& \makecell[c]{$69.97\pm2.03$} & \makecell[c]{$68.55\pm2.65$} & \makecell[c]{$72.39\pm2.95$} & \makecell[c]{$71.39\pm3.33$}\\                     
\bottomrule[1pt]
\end{tabular}
\label{tab:dis}
\end{table*}

\subsubsection{Ablative Experiments on LKCC Module}
\textbf{(1) The Impact of Prompt Number} The number of prompts is related to the quality of the global classifier. To study its impact, we only equip FedBM with the LKCC module and adjust the proportion of prompts to observe the performance change on the OCT-8 dataset.  As shown in Table~\ref{tab:ablation_prompt_number}, our method with different number of prompts presents different performance. Reducing the number of prompts probably leads to a decline in performance. When all prompts are used to construct local classifiers, FedBM achieves the highest performance. The experiment results demonstrate the importance of the number of prompts.

\textbf{(2) The Impact of Classifier Construction Method} To study the impact of the classifier construction method, we only equip FedBM with the LKCC module and change the method of classifier construction.  The vanilla FedAvg is regarded as the baseline. As shown in Table~\ref{tab:ablation_prompt_method}, freezing randomly-initialized local classifiers obtains better performance than the baseline. The results indicate that sharing a fixed classifier across clients is a feasible path to alleviate the classifier bias problem. Using averaged concept embedding as local classifiers outperforms the random initialization strategy, highlighting the effectiveness of linguistic knowledge. Notably, concept embedding distribution is superior to averaged concept embedding in different heterogeneous settings. This is because using embedding distribution as local classifiers can help the model capture the semantic diversification of image representations.

\begin{table*}[!t]
\caption{The performance of the proposed FedBM framework with different PLMs.} 

\vspace{-2mm}
\center
\renewcommand\arraystretch{1}
\setlength{\tabcolsep}{4pt}
\begin{tabular}{p{60pt}|p{60pt}|p{60pt}|p{60pt}|p{60pt}|p{60pt}}
\toprule
\multirow{2}{1.1cm}{\makecell[c]{Datasets} }&\multirow{2}{2cm}{\makecell[c]{PLMs} }      & \multicolumn{2}{l|}{\makecell[c]{$\beta=0.05$ }} & \multicolumn{2}{l}{\makecell[c]{$\beta=0.1$}} \\
\cmidrule(r){3-6}
&                 &    \makecell[c]{Accuracy (\%)}     &   \makecell[c]{F1-score (\%)}   &   \makecell[c]{Accuracy (\%)}        &  \makecell[c]{F1-score (\%)}  \\
\midrule
\multirow{4}{0cm}{\makecell[c]{OCT-8} }&\makecell[c]{Bert}&  \makecell[c]{$84.83\pm2.44$}  & \makecell[c]{$84.35\pm2.26$} & \makecell[c]{$87.82\pm2.06$} & \makecell[c]{$87.73\pm2.10$}  \\
&\makecell[c]{RoBERTa} &   \makecell[c]{$83.59\pm1.72$}  & \makecell[c]{$82.78\pm1.98$} & \makecell[c]{$\textbf{88.40}\pm2.48$} & \makecell[c]{$\textbf{88.35}\pm2.53$}  \\
&\makecell[c]{BiomedCLIP}& \makecell[c]{\textbf{$\textbf{84.84}\pm2.77$}}         &    \makecell[c]{\textbf{$\textbf{84.55}\pm2.78$}}         &     \makecell[c]{\textbf{$88.32\pm1.49$}}         &    \makecell[c]{\textbf{$88.16\pm1.44$}}         \\
\midrule
\multirow{4}{0cm}{\makecell[c]{Kvasir-v2} }&\makecell[c]{Bert}&  \makecell[c]{$69.48\pm3.25$}  & \makecell[c]{$67.64\pm4.42$} & \makecell[c]{$73.83\pm1.81$} & \makecell[c]{$73.19\pm1.58$}  \\
&\makecell[c]{RoBERTa} &  \makecell[c]{$70.73\pm1.58$}  & \makecell[c]{$\textbf{69.99}\pm1.69$} & \makecell[c]{$73.25\pm2.20$} & \makecell[c]{$71.52\pm3.51$}  \\
&\makecell[c]{BiomedCLIP}&  \makecell[c]{\textbf{$\textbf{71.31}\pm2.81$}}         &    \makecell[c]{\textbf{$69.51\pm4.11$}}         &     \makecell[c]{$\textbf{74.14}\pm 1.89$}       &    \makecell[c]{\textbf{$\textbf{73.75}\pm1.79$}}         \\
 \bottomrule
\end{tabular}
\label{tab:ablation_plms}
\vspace{-1mm}
\end{table*}

\subsubsection{Ablative Experiments on CGDE Module}
\textbf{(1) Evaluation of Key Components} Concept conditions, $\mathcal{L}_{div}$ and $\mathcal{L}_{dis}$ are the key components of the CGDE module. We remove them individually to observe the performance of our method on OCT-8 and Kvasir-v2 datasets. As shown in Table~\ref{tab:ablation_CGDE}, our method obtains the worst performance if we do use Gaussian noises instead of concept conditions on two datasets. Moreover, both removing $\mathcal{L}_{div}$ and $\mathcal{L}_{dis}$ result in performance degradation of the proposed method. The best performance is achieved by our method possessing three components simultaneously. Therefore, these experimental results confirm the importance of these components for the CGDE module.

\begin{figure}[!t]	
	\begin{center}
		\subfigure[OCT-8]
		{\includegraphics[width=0.93\columnwidth]{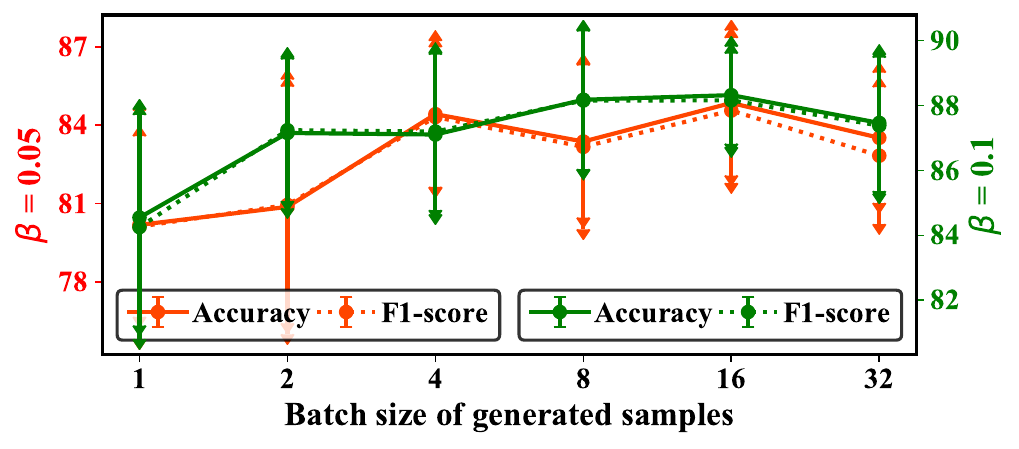}\label{fig:batchsize_generatedsamples_OCT-8}}
		\subfigure[Kvasir-v2]
		{\includegraphics[width=0.93\columnwidth]{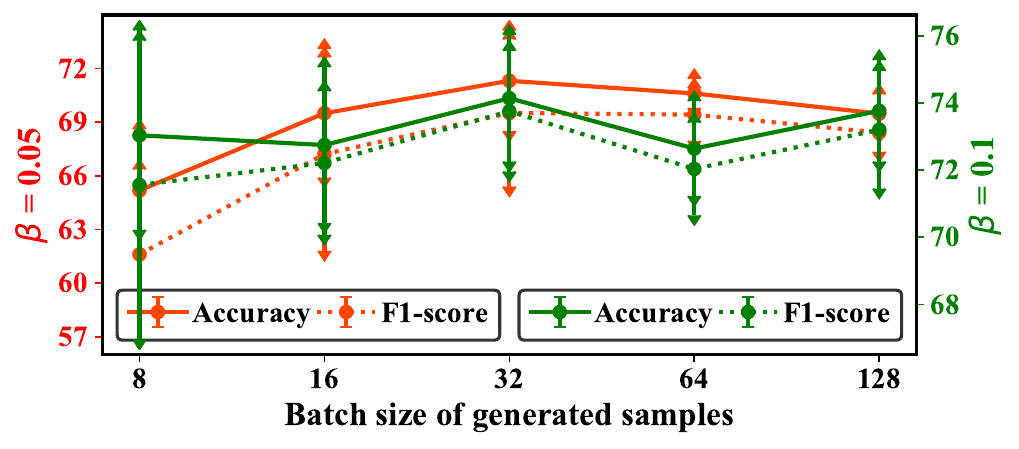}\label{fig:batchsize_generatedsamples_Kvasir-v2}}
	\end{center}
	\caption{The performance of our method with different batch sizes of generated samples on OCT-C8 and Kvasir-v2  datasets.}
    \label{fig:batchsize_generatedsamples}
\end{figure}

\textbf{(2) The Impact of Batch Size of Generated Samples} In Eq.~(\ref{eq11}), the larger batch size of generated samples indicates the stronger constraint on local updates. We fix the batch size of the original local data and compare the performance of FedBM with various batch sizes of generated samples. As presented in Fig.~\ref{fig:batchsize_generatedsamples_OCT-8}, FedBM obtains the lowest performance when the batch size of generated samples is $1$. As the batch size increases from 1 to 16 on OCT-8, the performance of FedBM exhibits an increasing trend both in Accuracy and F1-score for different $\beta$. In Fig.~\ref{fig:batchsize_generatedsamples_Kvasir-v2}, when the batch size of generated samples increases from 8 to 128, the performance of FedBM first rises to the highest point (the batch size is 32) and then shows a decreasing trend on Kvasir-v2. From these results, we can find that the constraint from a small batch size of generated samples is too weak to calibrate local updates, while a too-large batch size causes overcorrection of local updates.

\subsubsection{The Impact of Hyperparameters $\lambda_{div}$ and $\lambda_{dis}$}
$\lambda_{div}$ and $\lambda_{dis}$ in Eq.~\ref{eq10} are two important hyper-parameters of FedBM. The former controls the diversity of generated samples and the latter improve the training stability of the generator. To investigate their impact, we adjust $\lambda_{div} \in [0.1, 1, 10]$ and $\lambda_{dis} \in [0.1, 1, 10]$ to observe the performance of FedBM. As shown in Table~\ref{tab:div}, FedBM achieves the best performance on two datasets when $\lambda_{div}$ is set to 1. Therefore, we fix $\lambda_{div} = 1$ for all experiments.  In Table~\ref{tab:dis}, the highest performance is observed when $\lambda_{dis}$ is set to $0.1$ or $1$, so we present the best performance by selecting $\lambda_{dis} $ from $[0.1, 1]$ in all experiments.

\subsubsection{The Impact of Different PLMs}
To investigate the effect of pre-trained large language models, we equip the proposed FedBM framework with different PLMs, including Bert~\cite{devlin2018bert}, RoBERTa~\cite{liu2019roberta}, and the text encoder of BiomedCLIP~\cite{BiomedCLIP}. As shown in Table~\ref{tab:ablation_plms}, FedBM with different PLMs dose not present significant performance differences. Overall, BiomedCLIP yields better performance on two datasets compared with Bert and RoBERTa. The core reason may be that the text encoder of BiomedCLIP is trained on a medical text corpus, while Bert and RoBERTa are trained on text corpora that partially contain medical data.

\subsubsection{The Impact of Client Number}
\begin{figure}[!t]	
    \vspace{-1mm}
	\begin{center}
		\subfigure[Accuracy (OCT-C8)]
		{\includegraphics[width=0.49\columnwidth]{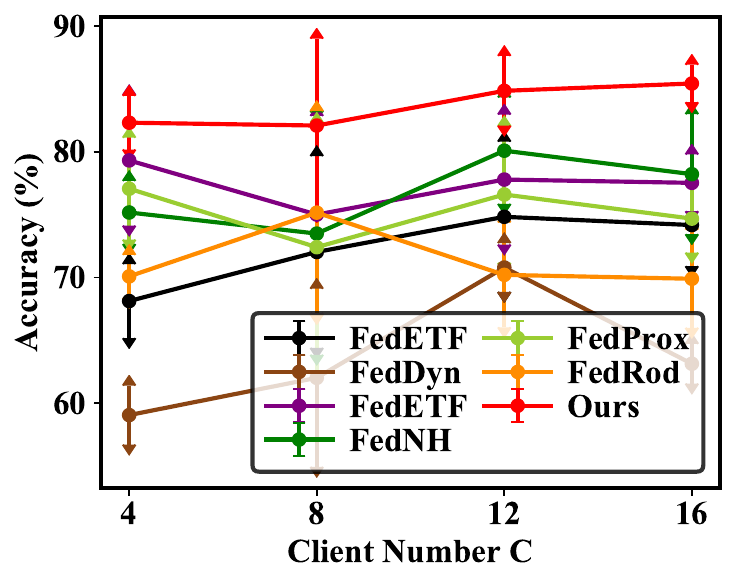}\label{fig:client_number_oct_acc}}
		\subfigure[F1-score (OCT-C8)]
		{\includegraphics[width=0.49\columnwidth]{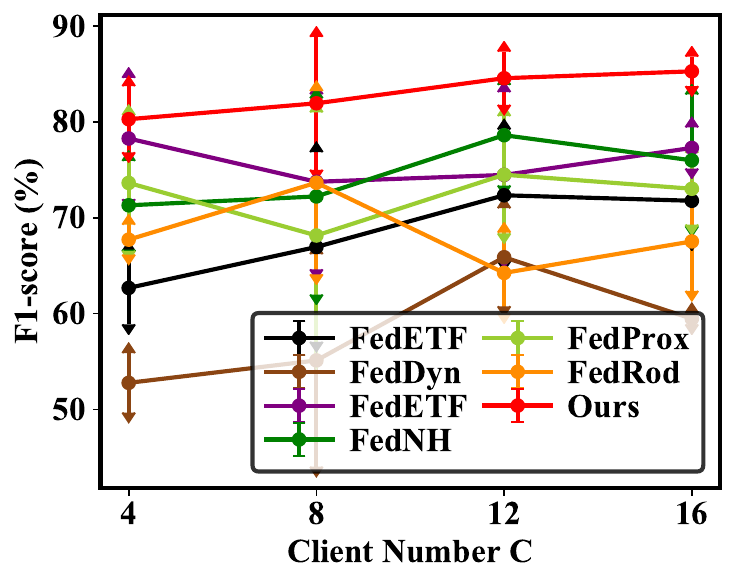}\label{fig:client_number_oct_f1}}
		\subfigure[Accuracy (Kvasir-v2)]
		{\includegraphics[width=0.49\columnwidth]{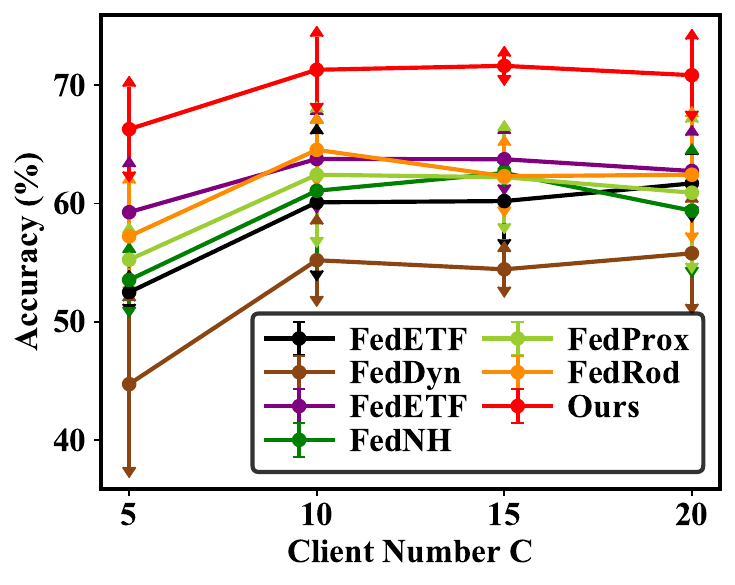}\label{fig:client_number_wce_acc}}
		\subfigure[F1-score (Kvasir-v2)]
		{\includegraphics[width=0.49\columnwidth]{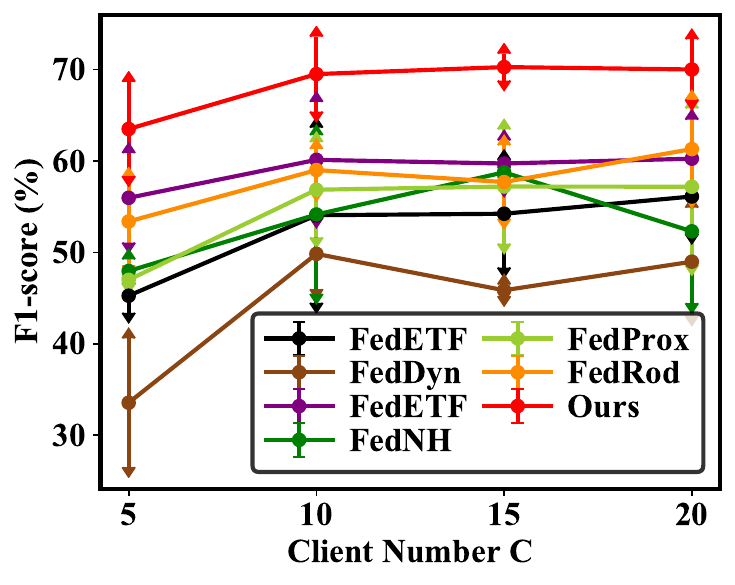}\label{fig:client_number_wce_f1}}
	\end{center}
	\vspace{-2mm}
	\caption{The performance of our method with different client numbers on Kvasir-v2 and OCT-C8 datasets.}
\label{fig:client_number}
	\vspace{-1mm}
\end{figure}

To compare the performance of different methods across different numbers of clients, we set the data heterogeneity $\beta = 0.05$ and divide the training data of OCT-8 and Kvasir-v2 datasets into $C$ clients, respectively. In Fig.~\ref{fig:client_number_oct_acc} and Fig.~\ref{fig:client_number_oct_f1}, the performance of previous methods display sharp fluctuations with the client number on OCT-8 dataset. By comparison, FedBM presents a more stable performance trend than these approaches. For Kvasir-v2 dataset, except for FedDyn and FedProx, although the performance of other existing methods is steady with respect to the number of clients, these methods achieve the limited performance. The proposed FedBM framework significantly surpasses all existing approaches for any client number. The experimental results can prove that the proposed FedBM is more robust against client numbers than existing methods.

\section{Conclusion}
\label{sec:conclusion}
In this paper, we propose a Federated Bias eliMinating (FedBM) framework to solve local learning bias problem in heterogeneous federated learning, which contains Linguistic Knowledge-based Classifier Construction (LKCC) and Concept-guided Global Distribution Estimation (CGDE).
LKCC can remove classifier bias by exploiting class concepts and pre-trained language models (PLMs) to construct a high-quality global classifier. CGDE is able to get rid of the learning bias of local feature extractors. It is based on probabilistic concept embeddings to learn a conditional generator. The generator is shared by all clients and produces pseudo data to calibrate updates of local feature extractors. 
The experimental results on five public datasets show the superior performance of FedBM in contrast to state-of-the-art methods under different heterogeneous settings. Extensive ablation experiments prove the effectiveness of submodules of FedBM. 

The proposed FedBM has achieved promising performance on various medical tasks, yet there are several limitations: (1) Although current PLMs are trained on large-scale datasets and show the strong generalization ability, some extreme cases may exist, such as highly similar concepts and open classes, for which our method is not suitable; (2) Theoretically, more diverse prompts can enable the proposed method to achieve the better performance. Our work is expected to provide a new respective to the research community for addressing data heterogeneity. The number of prompts in our experiments is not necessarily optimal. Both how to obtain diverse prompts and how to select the optimal number of prompts are two open directions worth exploring in the future; (3) In the proposed method, the generator is trained using the global model obtained by averaging the models from participating clients. Although there are not techniques available to recover the user-level privacy information for the generator, the copyright of client data may be infringed since other participants can use the generator to produce data for unintended purposes. A feasible solution to this problem is to train a generator that produces intermediate-layer feature maps rather than images.

\center\section*{Appendix}
We present the theoretical derivation of $\overline{\mathcal{L}}^\infty_{align}$ in Eq.~(\ref{eq4}).
\label{sec:appendix}
\begin{equation}
\footnotesize
\begin{aligned}
\small
\mathcal{L}^\infty_{align}&= \frac{1}{N_c}\sum^{N_c}_{i=1} \mathbb{E}_{\bm{e}^{(\textbf{y}_i)} \thicksim \mathcal{N}^{(\textbf{y}_i)}}
\left( -\log \frac{e^{\tau\textbf{h}_i\bm{e}^{(\textbf{y}_i)}}}{e^{\tau\textbf{h}_i\bm{e}^{(\textbf{y}_i)}} +\sum^{K}_{k\neq \textbf{y}_i}
\mathbb{E}_{\bm{e}^{(k)} \thicksim \mathcal{N}^{(k)}}
e^{\tau\textbf{h}_i\bm{e}^{(k)}}} \right) \\
&=\frac{1}{N_c}\sum^{N_c}_{i=1}\mathbb{E}_{\bm{e}^{(\textbf{y}_i)} \thicksim \mathcal{N}^{(\textbf{y}_i)}} \left(\log(e^{\tau\textbf{h}_i\bm{e}^{(\textbf{y}_i)}} + \sum^{K}_{k\neq \textbf{y}_i} \mathbb{E}_{\bm{e}^{(k)} \thicksim \mathcal{N}^{(k)}} e^{\tau\textbf{h}_i\bm{e}^{(k)}}) \right) \\
&\ \ \ \ -\frac{1}{N_c}\sum^{N_c}_{i=1}\mathbb{E}_{\bm{e}^{(\textbf{y}_i)} \thicksim \mathcal{N}^{(\textbf{y}_i)}} \left( \log(e^{\tau\textbf{h}_i\bm{e}^{(\textbf{y}_i)}})       \right) \\
&~~~~~//\textrm{using the Jensen's inequality:}\ \mathbb{E}[\log(X)]\leq \log(\mathbb{E}X)   \\
&\leq \frac{1}{N_c}\sum^{N_c}_{i=1} \left[ \log(\mathbb{E}_{\bm{e}^{(\textbf{y}_i)} \thicksim \mathcal{N}^{(\textbf{y}_i)}} \left(e^{\tau\textbf{h}_i\bm{e}^{(\textbf{y}_i)}} +  \sum^{K}_{k\neq \textbf{y}_i} \mathbb{E}_{\bm{e}^{(k)} \thicksim \mathcal{N}^{(k)}} e^{\tau\textbf{h}_i\bm{e}^{(k)}} \right)  \right]  \\
&\ \ \ \ - \frac{1}{N_c}\sum^{N_c}_{i=1} \left[  \mathbb{E}_{\bm{e}^{(\textbf{y}_i)} \thicksim \mathcal{N}^{(\textbf{y}_i)}}\log (e^{\tau\textbf{h}_i\bm{e}^{(\textbf{y}_i)}})  \right]  \\
&=\frac{1}{N_c}\sum^{N_c}_{i=1}\left[ \log(\sum_{k=1}^K \mathbb{E}_{\bm{e}^{(k)} \thicksim \mathcal{N}^{(k)}} e^{\tau\textbf{h}_i\bm{e}^{(k)}} ) - \tau\textbf{h}_i\bm{\mu}_{(\textbf{y}_i)}\right]  \\
&~~~~//\textrm{using the moment generation function for Gaussian} \\
&~~~~ \textrm{variable}\ X: \mathbb{E}[e^{\textbf{h}X}]=e^{\textbf{h}\bm{\mu}+\frac{1}{2}\textbf{h}^2\bm{\Sigma}}  \\
\end{aligned}
\label{eq12}
\nonumber
\end{equation}

\begin{equation}
\footnotesize
\begin{aligned}
\small
 \ \ \ \ \ \ \ \ &=\frac{1}{N_c}\sum^{N_c}_{i=1}\left[ \log(\sum_{k=1}^K e^{\tau\textbf{h}_i\bm{\mu}_k+ \frac{1}{2}\tau^2\textbf{h}_i^2\bm{\Sigma}_k}) - \tau\textbf{h}_i\bm{\mu}_{(\textbf{y}_i)}
\right]  \\
&~~~~//\textrm{Let}\ \mathcal{F}(\textbf{h},y) = \tau\textbf{h}\bm{\mu}_{(y)}+ \frac{1}{2}\tau^2\textbf{h}^2\bm{\Sigma}_{(y)} \\
&=\frac{1}{N_c}\sum^{N_c}_{i=1}\left[ \log(\sum_{k=1}^K e^{\tau\textbf{h}_i\bm{\mu}_k+ \frac{1}{2}\tau^2\textbf{h}_i^2\bm{\Sigma}_k}) -\mathcal{F}(\textbf{h}_i,\textbf{y}_i) +\mathcal{F}(\textbf{h}_i,\textbf{y}_i)
\right]  \\
& \ \ \ \ -\frac{1}{N_c}\sum^{N_c}_{i=1}\left[\tau\textbf{h}_i\bm{\mu}_{(\textbf{y}_i)}
\right]  \\
&=\frac{1}{N_c}\sum^{N_c}_{i=1}\left[-\log\frac{e^{\mathcal{F}(\textbf{h}_i,\textbf{y}_i)}}{\sum^K_{k=1}e^{\mathcal{F}(\textbf{h}_i,k)}}+\mathcal{F}(\textbf{h}_i,\textbf{y}_i) - \tau\textbf{h}_i\bm{\mu}_{(\textbf{y}_i)}\right] \\
&=\frac{1}{N_c}\sum^{N_c}_{i=1}\left[-\log\frac{e^{\mathcal{F}(\textbf{h}_i,\textbf{y}_i)}}{\sum^K_{k=1}e^{\mathcal{F}(\textbf{h}_i,k)}}+ \frac{1}{2}\tau^2\textbf{h}_i^2\bm{\Sigma}_{(\textbf{y}_i)}  \right] \\
&=\overline{\mathcal{L}}^\infty_{align}.
\end{aligned}
\label{eq12}
\nonumber
\end{equation}

\bibliographystyle{IEEEtran}
\bibliography{bare_jrnl}



%

\end{document}